\documentclass[10pt,compsoc,onecolumn]{IEEEtran}

\usepackage{times}
\usepackage{amsmath,amssymb}
\usepackage[applemac]{inputenc} 
\usepackage{ae,aecompl}
\usepackage{graphicx}
\usepackage{epsfig,alltt}
\usepackage{dsfont}
\usepackage{rotating}
\usepackage{cite}  
\usepackage{algorithmic}
\usepackage{algorithm}
\usepackage{array}
\usepackage{indentfirst}
\usepackage{subfigure}

\begin{document}

\title{Initialization Free Graph Based Clustering}

\author{Laurent~Galluccio,
	 Olivier~Michel,
	 Pierre~Comon,
	 Eric~Slezak,
	 and~Alfred~O.~Hero
\IEEEcompsocitemizethanks{\IEEEcompsocthanksitem L. Galluccio and P. Comon are with the I3S UMR 6070 CNRS, University of Nice Sophia Antipolis, 2000 route des Lucioles, 06903 Sophia Antipolis Cedex, France.
E-mail: \{gallucci,pcomon\}@i3s.unice.fr
\IEEEcompsocthanksitem O. Michel is with the Gipsa-Lab UMR 5216, 961 rue de la Houille Blanche - BP 46 - 38402 Saint Martin d  Heres Cedex, France. E-mail: olivier.michel@gipsa-lab.inpg.fr. 
\IEEEcompsocthanksitem E. Slezak is with the Nice Côte d'Azur Observatory
Boulevard de l'Observatoire - B.P. 4229 - 06304 - Nice Cedex 4, France.
E-mail: Eric.Slezak@oca.eu
\IEEEcompsocthanksitem A. O. Hero is with the Department of Electrical Engineering and Computer Science, University of Michigan, 1301 Beal Avenue, Ann Arbor MI 48109-2122, USA.
E-mail: hero@umich.edu.}}

\markboth{I3S Laboratory Internal Report, ISRN I3S/RR-2009-08-Fr, May 2009}%
{Galluccio \MakeLowercase{\textit{et al.}}: Initialization Free Graph Based Clustering}

\IEEEcompsoctitleabstractindextext{%
\begin{abstract}
This paper proposes an original approach to cluster multi-component data sets, including an estimation of the number of clusters. From the construction of a minimal spanning tree with Prim's algorithm, and the assumption that the vertices are approximately distributed according to a Poisson distribution, the number of clusters is estimated by thresholding the Prim's trajectory. The corresponding cluster centroids are then computed in order to initialize the generalized Lloyd's algorithm, also known as $K$-means, which allows to circumvent initialization problems. Some results are derived for evaluating the false positive rate of our cluster detection algorithm, with the help of approximations relevant in Euclidean spaces. 
Metrics used for measuring similarity between multi-dimensional data points are based on symmetrical divergences. The use of these informational divergences together with the proposed method leads to better results, compared to other clustering methods for the problem of  astrophysical data processing. 
Some applications of this method in the multi/hyper-spectral imagery domain to a satellite view of Paris and to an image of the Mars planet are also presented.
In order to demonstrate the usefulness of divergences in our problem, the method with informational divergence as similarity measure is compared with the same method using classical metrics. In the astrophysics application, we also compare the method with the spectral clustering algorithms. 
\end{abstract}

}

\maketitle
\IEEEdisplaynotcompsoctitleabstractindextext
\IEEEpeerreviewmaketitle
\section{Introduction}
\subsection{Motivation and contribution}
 \IEEEPARstart{O}{ne} of the recurrent problems in many fields of research, such as pattern recognition, machine learning or data mining, is data clustering. This is also a major subject of research in the remote sensing community, especially with the emergence of hyper-spectral sensors, since the amount of data available has grown significantly. \\
Clustering consists of partitioning a dataset into groups, such that points belonging to the same group have a high similarity, and points belonging to different groups have a low similarity. Similarity in this context means that data belonging to a cluster may share common features. \\ 

A new graph based approach is proposed for clustering large data sets ; a method for estimating the number of clusters present in the data and the approximate positions of their respective centers of mass is introduced in this contribution. More precisely, properties of minimal spanning tree (MST) and Prim's construction \cite{Prim57:bellst} are re-introduced and serve as a ground tool for the presented approaches. This approach is based on an idea developed in the context of bioinformatics by Olman et al. \cite{OlmaXX03:biocomputing}. It is proposed to record each iteration characteristics (namely which vertex is connected, and what is the length of the new edge). Exploiting further this idea, our method relates the $l$-dimensional probability density function of the data to Prim's curve. Prim's curve is interpreted as a \emph{one-dimensional} unfolded representation of the underlying data probability density function, exhibiting peaks and valleys. The number of clusters, which corresponds to the modes of the probability density function, is estimated by thresholding Prim's trajectory, in which sharply peaked modes are associated with deep valleys \cite{Stue03:JC}. \\
A Poisson model is introduced for the null hypothesis, in order to derive a test procedure that leads to determine an analytically optimal threshold for separating the clusters. The Poisson model also yields approximate convergence rates for the proposed technique.\\
An important part of this work is devoted to discuss the choice of the metrics or similarity measure, which is used in the multi-dimensional feature space, and required  for the computation of the similarity matrix. We propose novel information theoretic similarity metrics;  we illustrate that the performance of Prim's algorithm with information metrics significantly outperforms other clustering methods on several real world data sets. In the case of large data sets, an algorithmic solution is proposed, which allows to avoid the computation of huge similarity matrices based on data driven hierarchical classification. \\ 
Most of our discussion refers to K-means; two reasons explain that choice : first, K-means is one of the simplest and most popular approach for tackling clustering problems despite its sensitivity to initialization; second, K-means appears as a final step in the proposed algorithm, though other partitioning algorithms could be used. The present work is developed in the context of unsupervised clustering, and neither the underlying distributions, nor the number and locations of meaningful clusters are known; hence no prior information is required.
It must be emphasized that our use of K-means with Prim's MST is analogous to the use of K-means with Jordan \& Ng or Malik spectral clustering \cite{BachJ06:ML,NgJW01:nips,ShiM00:ieeepami}. Thus it is not simply a normalization but  a  new original hybrid approach that is presented here.

\subsection{Context and brief review of previous works}
The general clustering  problem has been studied following various approaches (see \cite{JainMF99:ACM} for an extended review). One can divide the various existing methods into two major classes: hierarchical and partitional clustering algorithms. 

Most hierarchical clustering algorithms are based on popular single-link or complete-link algorithms. These methods frequently suffer prohibitive computational time due to the construction of a dendogram on large datasets. 
Therefore, we are more interested in partitional methods, which are much more efficient. 
One approach is to consider that the dataset is drawn according to some probability distribution mixture, and the main goal is to estimate the number of distributions  and their parameters. Equivalently, one may consider the problem of detecting ``modes" or peaks within the data distribution.
Expectation Maximization (EM) \cite{DempLR77:jrsss} is a general method allowing to compute the maximum likelihood estimate of the parameters of an underlying distribution of a dataset, when these are incomplete or have missing values. This method suffers from major drawbacks. A distribution parametric model in each group needs to be known, even though it is in general assumed to be Gaussian. EM is highly sensitive to initial values of parameters, like most partitional clustering algorithms; it always converges, though not necessarily to a global optimum. 

One of the most popular  partitional clustering algorithm is the generalized Lloyd's algorithm, also referred to as $K$-means \cite{MacQ67:berkeley}. Given a finite number of clusters and an initialization of corresponding centers of mass, the $K$-means algorithm clusters the data in non-overlapping groups. It always converges (although not necessarily to the global optimum), by minimizing the intra-cluster variance. This algorithm, as well as the classical ISODATA algorithm \cite{BallH65:stanford}, belongs to the class of squared error algorithms, which aim at minimizing some objective function in order to cluster the data. As already mentioned, partitional clustering algorithms are highly dependent on initial parameters. In practice, for obtaining the best clustering results, the $K$-means algorithm is applied for many random initializations, and the best results are stored for later use.  

Another approach is spectral clustering, which is based on spectral graph theory \cite{Chun97:cbms}. These methods are based on the construction of a data similarity matrix and on the computation of eigenvalues of some graph Laplacian measures \cite{ShiM00:ieeepami,NgJW01:nips,BachJ06:ML}. Spectral clustering methods outperform single-cut clustering, which is not robust with respect to outliers. In fact, the latter takes into account the size of each cluster in determining the cut value. \\

Finding the right number of clusters in a dataset is a key issue, present in various fields of research. 
Several methods have been proposed to estimate this number, for example using statistical criteria like Akaike information criterion (AIC), Bayesian information criterion (BIC), minimum description length (MDL) \cite{Riss89:worldscientific}, Tibshirani's gap \cite{TibshWH01:jrss}, or indices such as Calinski \& Harabasz's index \cite{CaliH74:stat}. Most of these techniques are based on a clustering step with $K$-means, then by using these predefined criteria, clusters are merged or split. \\

More precisely,  in the vector quantization community, Bischof et al.\cite{BiscLS99:PAA} propose to use a concept related to algorithmic complexity, namely MDL, in order to reduce the number of clusters initially given (starting with a high value of $K$). \\
Pelleg and Moore \cite{PellM00:sanfrancisco} have developed a variant of $K$-means referred to as $X$-means, starting from a minimum number of clusters, then applying $K$-means. The cluster splitting procedure is based on the BIC criterion. \\
Hamerly and Elkan \cite{HameE03:nips} propose  to learn the integer $k$ required in $K$-means by assuming a gaussian mixture model. Starting with a small value of $k$, the algorithm splits the clusters whose distribution is not Gaussian. Between each statistical test, the $K$-means algorithm is applied to refine the solution.
Nevertheless, this algorithm referred to as $G$-means performs poorly in the presence of non-spherical or non-elliptical clusters. \\
 Tibshirani et al. \cite{TibshWH01:jrss} have defined the Gap statistic to determine the optimum number of clusters. Their method is based on the output of a clustering algorithm: $K$-means or hierarchical classification. The authors propose to compare the logarithm of the within-cluster dispersion to its expectation under a reference null-distribution. This method performs better if clusters are well separated and if the number of clusters is small.

Another parameter having a huge impact on the performance of partitional clustering algorithms is the location of initial cluster centers. 
That explains why refinement of initial conditions for the $K$-means algorithm has been widely studied during the last decade (see \cite{PenaLL99:prl} for a short comparative study of initialization methods for $K$-means). Since a ``wrong" initialization of this algorithm leads to a convergence to local minima, various methods have been developed in order to provide the best convergence possible (to a local minimum if the global minimum hasn't been reached). A good way to initialize properly this algorithm consists of placing each cluster centroid at the  modes of the joint probability density of the data. Motivated by this idea, Bradley and Fayyad \cite{BradF98:icml} propose a refinement of initial conditions of $K$-means near the modes of the distribution estimated, by the use of a recursive procedure executing $K$-means on small random sub-samples of the data. Likas et al. \cite{LikaVV03:PR} propose a new method based on a recursive partitioning of the data space into disjoint subspaces by using $k-d$ tree methods, and by defining the cutting hyperplane as the one perpendicular to the highest variance axis. 

Our concern is to derive an unsupervised clustering scheme acting on large multi-dimensional data sets.

\subsection{Organization of the paper} 
In Section \ref{section:2}, definitions and properties of MST are detailed; then, our approach is introduced based on the use of information extracted from the construction of a MST with  Prim's algorithm. 
Some analytical  results are derived  for evaluating the false alarm rate of our cluster detection algorithm, using a Poisson model for the distribution of the MST vertices in the data features multi-dimensional space. \\

The construction of a MST requires the computation of a similarity matrix, based on distances between all pairs of points among the dataset. This computation becomes prohibitive in the case of large datasets. In Section \ref{section:3}, in order to circumvent this problem, a method using a preconditioning data-driven hierarchical classification tree is developed, in order to decrease the computational time of the MST construction. With this goal, the assumption that the vertices are approximately distributed according to a Poisson distribution is made.

Since the partitioning problem is based on similarity between points, we address the problem of designing a measure of similarity between two multi-dimensional data points. Instead of using the popular Euclidean distance and the spectral angle mapper (SAM), which are a common measure used in hyper-spectral processing \cite{Kesh04:ieeegrs}, we propose to define the similarity as a measure of spectral variability between two probability density functions, where informational divergences are involved. In Section \ref{section:4}, new metrics and motivation for resorting to information based similarity measures are investigated.

In Section \ref{section:5}, we present some unsupervised clustering results obtained in the frame of three applications: the taxonomic classification of asteroids, the classification of objects in a multi-spectral satellite image and the classification of chemical species in a hyper-spectral image of the Mars planet.

\section{Automatic initialization} \label{section:2}
In this section, we briefly review the theory of MST and Prim's algorithm. Then, based on the construction of the MST, we present our original approach to initialize a partitioning algorithm by estimating the number of clusters and the location of their corresponding centroids. Example are derived with the K-means algorithm.

\subsection{Formulation}
Let $V$ be a set of $N$ data points in $\mathbb{R}^L$. Each data point (actually a vector in $\mathbb{R}^L$) is considered as a vertex in a graph. Hence, the goal is to partition $V$ into $k$ classes. Denote $C : (C_1, \ldots, C_k)$ the set of clusters, and let $\mu : (\mu_1, \ldots, \mu_k)$ be the set of corresponding centroids. Let us emphasize that no prior information is introduced. The present work is developed in the context of unsupervised clustering, and neither the underlying distributions, nor the number and locations of meaningful clusters are known. 

\subsection{Minimum Spanning Tree and Prim's algorithm}
Let $G=(V,E)$ be an undirected graph where $V = (v_1, \ldots, v_N)$ is the set of $N$ vertices and $E$ denotes the set of edges. The length of an edge measures the similarity between two vertices, and depends on the choice of the metric.
The graphs considered herein are \emph{trees}. They are connected, which means that every vertex is connected to at least one other, and acyclic (i.e., there is no loop).
\medskip

A spanning tree of $G$ is a tree $T$  passing through every vertex of $G$. 
The power-weighted length of the tree $T$ is the sum of all edge lengths raised to a power $\gamma \in (0,L)$, denoted by: $\sum_{e\in T} |e|^{\gamma}$. The minimal spanning tree is the tree that has the minimal length over all spanning trees:
\begin{equation}
{\cal L}(V) = \min_{T} \sum_{e\in T} |e|^{\gamma}\label{L_MST}
\end{equation}

\medskip

Several algorithms have been developed in order to construct a MST. 
One can cite for instance the most popular ones: Boruvka's, Kruskal's and Prim's (see \cite{GrahH85:hc} for a state of art of various MST's algorithms). 
It is Prim's algorithm \cite{Prim57:bellst} that will be the primary focus of this paper. 

Let $T_i$ be the graph partially connected at iteration $i-1$, hence $i$ vertices are connected to the MST. At the $i$th iteration, one non-connected vertex, say $v_i$, and one connected vertex of $T_i$ are selected, so that the similarity measure between them is minimized; $T_i$ becomes $T_{i+1}$ with this new vertex $v_i$ and the associated edge of minimal length. This operation is repeated until no unconnected vertex remains. In other words, $N-1$ iterations are required to construct the complete MST from a set of $N$ points.

The choice of the initial vertex of the construction of the graph is not important because the complete MST is unique (if there are no ties in the pairwise distances). Only the descriptor is a function of the initial vertex, but all descriptors will be identical up to a permutation of pairs (edge, vertex). The dissimilarity matrix is given as an input argument to the MST algorithm, and since the graph is undirected, this matrix has to be symmetric. The importance of the choice of the metric provided to the MST computing routine is discussed in Section \ref{section:4}.

To sum up, the graph obtained this way is acyclic (no loop), unique (that is, independent of the initial point of construction) and of minimal length.

\subsection{Prim's Trajectory}
Many graph-theoretic clustering algorithms are based on edge cuts of a MST built through the set of data \cite{JainMF99:ACM}. A simple way to segment the graph into $k$ clusters is to remove the $k-1$ largest edges of the graph. This method, similar to single-linkage clustering \cite{GoweR69:AS}, is known to be unstable, mainly when the data contain outliers, or possibly when they are corrupted by noise. 
\medskip

\begin{figure}[htb]
\centering
\includegraphics[width=9cm,height=9.5cm]{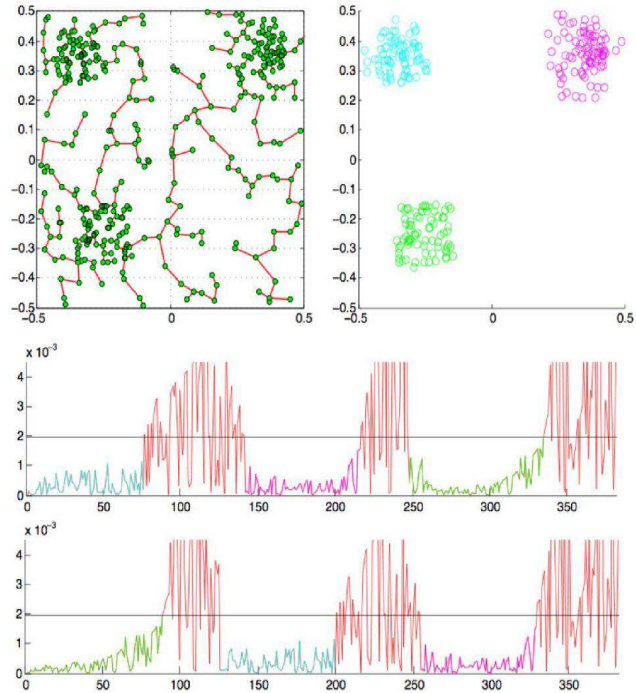} 
\caption{\small{Typical example: (top left) construction of a MST, (top right) extraction of the clusters, (bottom) Prim's trajectories. We have rooted the MST from two random initial vertices and recorded the descriptors information. Although the MST is unique, the Prim's trajectory doesn't follow this condition; nevertheless this property has no effect on the extraction of clusters. Then, these information, number of clusters and corresponding centroids (computing on the extracted clusters), are used to initialize a classical algorithm: for example $K$-means.}} \label{MST-prim}
\end{figure}

Denote $g(i)=|e_i|$, the length of a new edge built by Prim's algorithm at iteration $i$; $[g(i), i=1\ldots N ]$ is referred to as \emph{Prim's trajectory}. Function $g$ allows us to ``unfold'' the MST over points in $L$ dimensions into a one-dimensional function (see Fig.~\ref{MST-prim}).

Though a Prim's trajectory is closely related to the way the MST is constructed, the function $g$ is not uniquely associated with a MST. While the Prim trajectory $g$ depends on the choice of the initial vertex, the MST does not.

In Fig.~\ref{MST-prim}, this property is illustrated: the same MST is built from two different Prim's trajectories. The property of non-uniqueness of Prim's trajectories does not affect our ability to identify and extract.
The occurrence of valleys in the curve hence corresponds to the regions of the MST where the Prim algorithm transitions between clusters. It allows to identify the main modes of the probability density function, each mode being associated with a cluster.
Actually this concept of using a one-dimensional function representing distances between connected vertices in order to cluster the data points has already been proposed by Slagle et al. \cite{SlagCL74:PR}. They propose to build a short spanning graph, plot on a straight line vertices of the graph where distances between points are respected and threshold this one dimensional function in order to reveal clusters. We can also cite the works of Stuetzle \cite{Stue03:JC} who proposes to analyse the MST of a sample to find modes of a density.

\medskip

The choice of the threshold to be applied to $g(i)$ is an important parameter, since it determines the main modes of the probability density function. Michel et al. \cite{MichBR05:toulouse} have proposed to give the same cluster label to all vertices connected sequentially, as long as the function $g(i)$ stays below the threshold and proposed an empirical solution for setting the threshold. 
One possibility is to arbitrarily set the threshold to the standard deviation of the edge length $\epsilon = \sigma(e_i) $.

 For the sake of simplicity, the threshold has been chosen to be constant over the entire Prim's trajectory, but it could vary from one cluster to another. 
This leads to a good refinement of initial conditions of $K$-means and a fast convergence, since it starts at centers of mass of previously detected clusters.

\medskip

However, the question remains to determine the critical number of points  that should be considered to build a cluster. Actually for a given realization, some vertices can be gathered within a small neighborhood, even in the case where the theoretical density does not exhibit any local maximum.  Alternatively, small clusters may correspond to noise effects and thus may not be relevant. Thus, the number of modes tends to be overestimated.
In the next section, we show how this critical number can be estimated using a Poisson model. 

\subsection{Estimation of the minimum number of points in a detected cluster}
The estimation of the minimum number of points in a detected cluster is formulated as a Neyman-Pearson detection problem. For each new connected vertex, a binary hypothesis test is performed. Under the null hypothesis, the most recent connected vertices do not belong to any 'mode' or cluster. Under the alternative hypothesis, the new connected vertex belongs to a mode of the density.

Let $v_i$ be the vertex connected at iteration $i$. 
Consider an $L$-dimensional space and a neighborhood of $v_i$, hereafter denoted $B_{(v_i,\epsilon)}$, with characteristic length $\epsilon$. We suppose that the vertices are approximately distributed according to a Poisson distribution,  with rate $\lambda \epsilon^L$. This is justified, since the latter is the limiting distribution of the binomial for a large number of points. Another justification of the Poisson approximation for the distribution of vertices in the case of the construction of a MST is presented by Steele \cite{Stee97:siam}. Both $\lambda$ and $B_{(v_i,\epsilon)}$ will be identified later. 

\begin{figure}[htb]
\centering
\includegraphics[width=70mm]{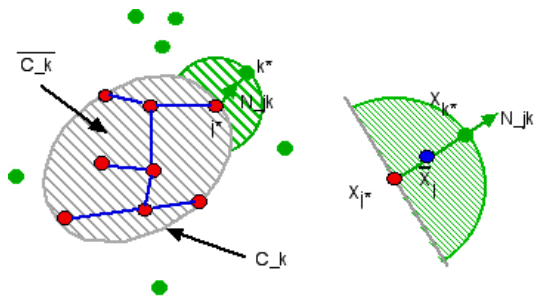}
\caption{\small{Left : $C_k$ denotes the contour of the support where connected vertices are found. The neighborhood which is considered for finding a vertex that could be connected to $v_i$ is shown. In the limit of large $N$, this latter neighborhood is the half sphere laying on the tangent (hyper)plane to $C_k$ (right). }}\label{Contour}
\end{figure}

We assume that under the hypothesis $H_0$, the density does not exhibit any mode:  the process is homogeneous  over its entire support $\cal V$. Hence we can assess that   $\lambda$ is constant over $\cal V$. The probability that at least one vertex is found in the neighborhood $B_{(v_i,\epsilon)}$ is given by:
\begin{equation}
F_{v_i}(\epsilon)=1-{\rm e}^{-\lambda\epsilon^L}.\label{equ:F1}
\end{equation}
In the context of the Prim construction of the MST, $F_{v_i}(\epsilon)$ is the probability to construct an edge of length less than $\epsilon$ when connecting a new vertex to $v_i$. It corresponds to the cumulative distribution function of the length of the shortest edge.
Considering the asymptotic case where $N$ is large, the neighborhood which must be considered here is the half sphere of radius $\epsilon$, as illustrated in Fig.~\ref{Contour}.

\medskip

Suppose that at least $k_0$ successive connections of length less than $\epsilon$ are required for considering that a cluster is detected. Here, parameter $\epsilon$ corresponds to the threshold value applied to Prim's trajectory. Under hypothesis $H_0$ and the assumption that all vertices are mutually independent, false alarm in the mode detection will arise for any occurrence of more than $k_0$ successive connections of length less than $\epsilon$. Therefore from (\ref{equ:F1}), the expression of the false alarm probability is given by:
\begin{equation}
P_{FA}(k_0, \epsilon)  = \left(1-{\rm e}^{-\lambda\epsilon^L}\right)^{k_0}.\label{equ:pfa}
\end{equation}
In the case where $L$-dimensional Euclidean spaces are considered, the volume of the half sphere of radius $\epsilon$ is  $B_L(\epsilon)=\frac{1}{2}C_L\epsilon^L$, where $C_L$ stands for the volume of the unit ball in dimension $L$: $C_L=\frac{2\pi^{L/2}}{L\Gamma(L/2)}$ (where $\Gamma$ is the Gamma function). \\

\begin{figure}[htb]
\centering 
\includegraphics[width=9cm,height=6cm]{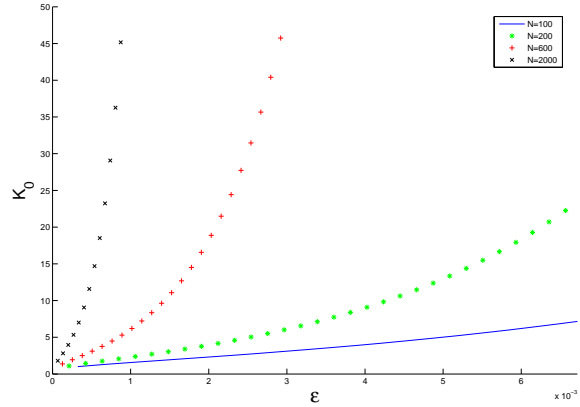} 
\caption{\small{Minimum number of points in a detected cluster as a function of the threshold value $\epsilon$ from sets of uniformly distributed vertices over $[0, 1]^2$, and $N=100, 200, 600,2000$ vertices respectively.}}\label{figure_k0}
\end{figure}

In this framework, under $H_0$, consider the radius $\epsilon_0$ of a sphere covering the set of all vertices; $\lambda$ is identified by the set of equations: 
\begin{equation}
\begin{array}{ccc} \left\{ \begin{array}{ll} {\cal V} &= C_L \epsilon_0^L\\ \lambda\epsilon_0^L&=N\end{array}\right. & 
\Longrightarrow &  \lambda = C_L\frac{N}{{\cal V}}\end{array}\label{equ:lambda}
\end{equation}
Finally, we get from (\ref{equ:pfa}) and (\ref{equ:lambda})
\begin{equation} P_{FA}(k_0, \epsilon)  = \left(1-{\rm e}^{-C_L\epsilon^L\frac{N}{{\cal V}}}\right)^{k_0} \end{equation}
This formula determines the relationship between $k_0$ (minimum number of vertices that form a cluster) and the threshold value $\epsilon$ in the framework of a Neyman-Pearson test (Fig.~\ref{figure_k0}). 
\begin{equation}
k_0 = \frac{\log(P_{FA})}{\log\left(1- \exp \left({-C_L \epsilon^{L} \frac{N}{{\cal V}}}\right)\right)}
\end{equation}

\medskip

From (\ref{equ:F1}) we derive the probability that at least $k$ successive connections are of length less than $\epsilon$.
\begin{equation}
P^{*}_{k,\epsilon}= \prod_{i=1}^k F_{v_i}(\epsilon) = (1- e^{-\lambda \epsilon^L})^k  \label{equ:pk}
\end{equation}
Let $P_{k,\epsilon}$ be the probability that exactly $k$ vertices are connected with  edge lengths less than $\epsilon$ but the next edge built is larger than $\epsilon$ : 
$$ 
P_{k,\epsilon}=P^{*}_{k,\epsilon}(1-F_{v_i}(\epsilon))
$$
From (\ref{equ:pk}) and (\ref{equ:F1}), we obtain the following expression
\begin{equation}
P_{k,\epsilon} =\left(1-{\rm e}^{-\lambda\epsilon^L}\right)^k{\rm e}^{-\lambda\epsilon^L}\label{equ:pkeps}
\end{equation}

The asymptotic behavior of (\ref{equ:pkeps}) when varying $\epsilon$ tends to validate this expression of probability in our problem.
$$ 
\begin{array}{c l}
\underset{\epsilon \rightarrow \infty}{\lim} P_{k,\epsilon} = 0& \mbox{for a given }k \mbox{ value},\\
\underset{\epsilon \rightarrow 0}{\lim} P_{k,\epsilon} = 0 & \mbox{except the case where } k=0.\\
\end{array}
$$ 

At a given threshold value $\epsilon$, we verify that $P_{k,\epsilon}$ is indeed a probability density:
$$\sum_{k=0}^{\infty} P_{k,\epsilon} = e^{-\lambda \epsilon^L}  \sum_{k=0}^{\infty} (1-e^{-\lambda \epsilon^L})^k = 1 $$ 

\medskip

Similarly, under $H_0$, one is able to estimate the average number of vertices included in ``false alarm clusters''.
At a given threshold value, the average number of points of a false alarm cluster detected is defined by:
\begin{eqnarray}
<k> &= &\sum_{k=1}^{\infty} (k+1) P_{k,\epsilon}\nonumber \\
 & = & e^{-\lambda \epsilon^L}  \sum_{k=1}^{\infty} k (1-e^{-\lambda \epsilon^L})^k + 1 - e^{-\lambda \epsilon^L}\nonumber \\
  & = &(1-e^{-\lambda \epsilon^L} ) + \frac{(1- e^{-\lambda \epsilon^L})}{ e^{-\lambda \epsilon^L}}\nonumber \\
 & =&2 \sinh(\lambda \epsilon^{L/2}) =2 \sinh \left(C_L\frac{N}{{\cal V}}\epsilon^{L/2}\right)\label{average_k}
 \end{eqnarray} 
In Fig.~\ref{figure_k0}, we compare the behavior of the minimum number of points required in a cluster detected $k_0$ when the threshold value varies and for different sets of uniformly distributed points. For small sets of points, a little variation of the threshold value has no impact on the number $k_0$. As the set of points increases, the estimation of the parameter $k_0$ is highly dependent on $\epsilon$. 
 
\begin{figure}[htb]
\centering
\includegraphics[width=9cm,height=6cm]{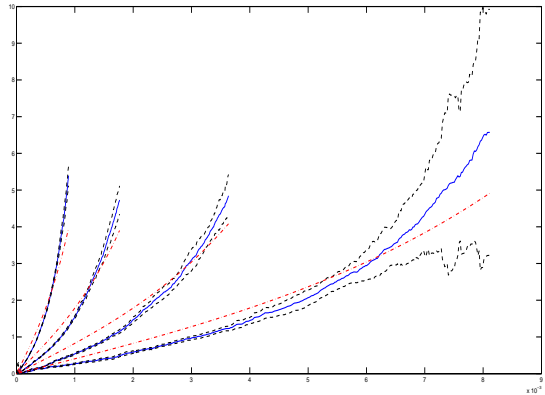} 
\caption{\small{Average size of false alarm detected cluster from sets of uniformly distributed vertices over $[0, 1]^2$, and $N=128, 256, 512,1024$ vertices respectively (curves from the right to the left). Theoretical curve (red) and numerical simulation (blue: mean, dotted black: $\pm$ standard deviation).}}
\label{figure_simul}
\end{figure}

In Fig.~\ref{figure_simul}, we have plotted the function linking the threshold and the average number of points of a false alarm cluster detected, obtained through numerical simulations and theoretical relations (\ref{average_k}) on datasets of different sizes. 
Though these calculations were derived under simplifying assumptions, the obtained results are in good agreement with simulations, as illustrated in Fig.~\ref{figure_simul}. 

The above results apply to the case that the edge length is measured by Euclidean distance. In Section \ref{section:4}, we develop alternative measures of edge length that are asymptotically equivalent to Euclidean distances in the limit of large N, when the vertices become close to each other.

\section{Implementation of MST's for large datasets}\label{section:3}
As shown above, Prim's algorithm requires a dissimilarity matrix as input parameter, which leads to the computation of $N\times (N-1)/2$ distance values between all pairs of points. This computation becomes prohibitive when the dataset is large. We propose a fast data hierarchical algorithm, allowing the calculation of dissimilarity only between small sets of neighboring vertices.
Accordingly, the method developed identifies the nearest neighbors of every vertex. 

Finding nearest neighbors of a given point in space within an efficient computation time has been widely addressed. One of the most famous search algorithm is the space partitioning technique based on a $k$ dimensional binary search tree, referred to as the $k-d$ tree \cite{FrieBF77:ACM}. The optimized $k-d$ tree method chooses the hyperplane passing through the median point perpendicular to the coordinate axis where the distribution exhibits the largest spread. 
Based on a search for nearest neighbors using the $k-d$ tree and priority queues to reduce these searches, Bentley and Friedman \cite{BentF78:ieeec} proposed several algorithms of MST construction.
Another space partitioning structure has been developed by Guttman \cite{Gutt84:ACM}: the R-tree. Although in the $k-d$ tree methods the partitioning element is an hyperplane, the R-tree uses hyper-rectangular regions. Both methods can hardly handle high dimensional data, since their complexity grow exponentially with the dimension. The same problem appears in other methods such as $v-p$ trees \cite{Yian93:philadelphia}, quad-trees \cite{FinkB74:informatica} or hB-trees \cite{LomeS90:ACM}.

The construction of the MST requires a sort the lengths of all possible edges. This operation requires $Comp.sort=O(N\log(N))$ logical operations (for example by using the ``quick-sort'' algorithm \cite{PresTvF07:Cambridge}). Thus the overall computational cost is mostly due to the computation of all the distances. 
This  becomes prohibitive for large datasets, since it would lead to a computational burden of $Comp.dist=O(L\,N^2/2)$ flops. 

This is avoided by using a preconditioning data-driven hierarchical classification tree, which could be learned from a small randomly chosen subset of $R$ data points.\footnote{The hierarchical tree requires $O(LR\log R)$ logical operations.} The classification tree allows to identify neighborhoods of each vertex by comparing the vertex coordinates along each of the $L$ dimensions to sets of thresholds. The size of the neighborhood can be set up in such a way that in the average, each neighbor\-hood contains about $M$ vertices. Consequently, the number of pairwise distances that need to be evaluated is of the order of $M^2/2$. This algorithm leads to a maximal computational burden expressed by $Comp.dist = O(NL\,M^2/2)$ flops and $Comp.sort= O(NM\log M)$. Assuming that $M$ is set such that $M^2\ll N$, the computational load is thus significantly lowered. In the next sections, constructions of MST using this algorithm will be referred to as `Nearest-Neighbor MSTs (NN MST). 

For each coordinate, a sample set of data is used in order to determine recursive median splitting; this leads to some binary hierarchical tree which counts $2^t$ cells (terminal leaves) if there are $t$ recursive iterations. Consequently, a new data vector is affected to a final leaf by applying $t$ successive thresholding operations. 

A natural criterion to evaluate the number $t$ of splits could be to compare the number of points within the cell to be split and the variance of the median estimate \cite{MoodGB74:McGrawHill}; let $N_{\Pi}$ be this number. 
In the present situation, some hypercells will be defined by considering median-based hierarchical structure over each axis. Therefore, a correct criterion to use would be $N_{\Pi}^L$. 

Let $\Pi$ be the hypercell containing nearest points of dimension $[\alpha_i,\beta_i]_{i=1}^L$. 
Let $V_{\Pi}^{j}$ be the $j$-th coordinates of $V$ falling in the terminal leaf $\Pi$. We assume that the elements of $V_{\Pi}^{j}$  are independent and identically distributed (i.i.d.) with a uniform density $f_{v^j|\Pi}$: $f_{v^j|\Pi}= 1/ (\beta_j - \alpha_j)$. The sample median $\hat{T}_{\Pi}^j$ is asymptotically distributed as a normal distribution \cite{MoodGB74:McGrawHill}: 
\begin{equation}
 \hat{T}_{\Pi}^j \sim \mathcal{N} \left(T_{\Pi}^j, \frac{1}{4N_{\Pi} (f_{v^j|\Pi}(T_{\Pi}^j))^2} \right) \label{equ:Testimate}
 \end{equation}
where $T_{\Pi}^j$ is the theoretical median: $T_{\Pi}^j = (\beta_i + \alpha_i) / 2$ and $N_{\Pi}$ is the number of vertices within $\Pi$. The sample medians $\hat{T}_{\Pi}^j$ are statistically independent.

Therefore, a natural criterion is to require that $N_{\Pi}$ is large enough so that the density  of $\hat{T}_{\Pi}^j$ has maximum mass inside the interval $[\alpha_i,\beta_i]$ (and hence a minimum mass outside): 
\begin{equation}
1 - \prod_{j=1}^{d} P(| \hat{T}_{\Pi}^j - T_{\Pi}^j | \leq  (\beta_j - \alpha_j) / 2 ) \leq \epsilon
\end{equation}
Under the Gaussian approximation (\ref{equ:Testimate}) becomes 
$$
1 - Pr(|Z| < \sqrt{N_{\Pi}})^L \leq \epsilon 
$$
where $Z$ is a standard normal random variable (zero mean and unit variance). 
$$
Pr(|Z| < \sqrt{N_{\Pi}} ) = erf(\sqrt{N_{\Pi}/2})
$$
where $erf$ is the error function: $erf(x)=\frac{2}{\sqrt{\pi}} \int_0^x e^{-t^2} dt$.\\

Thus, we can determine the following stopping criterion about the splitting cell procedure according to the number of points in the cell.
\begin{equation}
N_{\Pi}  \geq 2 \left[ erf^{-1}((1-\epsilon)^{1/L})\right]^2 
\end{equation}

Another problem must be also considered: once a vertex is associated with a cell (after $t$ logical tests), its closest neighbor is possibly in some adjacent cell, rather than in the cell under investigation. Because the median split over each axis leads to a simple logical structure of the overall classification tree, the hypercube formed by the cell and the set of its $3^L$ adjacent neighbors is easily computed and all similarities between pairs of vertices from this hypercell are considered.

In Table \ref{NNMST_Descriptors}, we describe the hierarchical structure developed by the recursive splitting algorithm. The structure for $L$ dimensions is obtained by stacking up every descriptors. The descriptor of each node is composed by $5$ components for each dimension. Then, at each splitting of a cell, two others columns are added (since this is a binary tree). Therefore, once this descriptor matrix is built, the nearest neighbor problem is reduced to a search in this matrix.  

\begin{table}[htb]
\centering
\caption{\small{NN MST Descriptors for the $\ell$th tree.}}\label{NNMST_Descriptors}

\begin{tabular}{l l}
\hline
\hline
$\mathbf{Output:}$&  average of the $\ell$th coordinate, $1\le\ell\le L$  (or for example \\
& variance of the distribution  of the points) \\
$\mathbf{Status:}$ & $0 \rightarrow$ this leaf has to be split \\
& $1$ $\rightarrow$ terminal leaf \\
& $10$ $\rightarrow$ empty leaf, or not enough points in the leaf \\
$\mathbf{Parent:}$& parent index in this descriptor table\\
$\mathbf{Child:}$& index of the $1^{\mbox{st}}$ child in the table (in the second 
 column \\&of this descriptors table, this parameter will contain the \\& index of the $2^{\mbox{nd}}$ child in the table)\\
$\mathbf{Depth:}$ &tree depth\\
$\mathbf{Card:}$& number of points presents in each node\\
\hline
\end{tabular}
\end{table}

\begin{algorithm}

\caption{Storage of the NN MST Descriptors}
\label{Algorithm2}
\begin{algorithmic}[1]

\REQUIRE sampled time series and embedding parameters
\ENSURE NN MST descriptors\\
{\small{\COMMENT{Growing algorithm building the descriptor matrix, described here in dimension $L$ possibly larger than 1}}}
\STATE Initialize $\Pi^0 =$ set of all state vectors
\WHILE{non-empty non-terminal leaves exist, at current depth $l$}
\FORALL{each cell $\Pi^l$}
\IF{$N_{\Pi^l} \geq 2 \left[ erf^{-1}((1-\epsilon)^{1/L})\right]^2$ vectors}
\STATE compute the splitting thresholds $T_{\Pi}^j, j=1,\ldots,L$.
\ELSE 
\STATE $\Pi^l$ is stored as a terminal leaf.
\ENDIF
\ENDFOR
\STATE $l=l+1$
\ENDWHILE
\end{algorithmic}
\end{algorithm}


The procedure for establishing the NN MST Descriptor matrix is summarized by Algorithm 1 detailed in Table \ref{Algorithm2}. This algorithm is to be executed $L$ times with $L=1$ instead of once in dimension $L$.
Obviously the computational cost of this algorithm depends on the number $N_{\Pi}$ previously defined.

\section{Choice of metrics}\label{section:4}
The distance measure between points plays a key role in clustering performance, in the sense that it characterizes similarity or dissimilarity between points.
Therefore, the use of different metrics will lead to different Prim's trajectories. Hence, we have to pick up the ``right" measure that better fits the data.

\noindent In this paper, we assume that data points are semi-definite positive.\\
Let $X=\{x_1,\ldots,x_L\}$ and $Y=\{y_1,\ldots,y_L\}$ be two feature vectors.
The most popular distance used to characterize similarities between two points is the Euclidean distance: 
$$d(X,Y)= \sqrt{ \sum_{i=1}^L (x_i - y_i)^2}$$

Though this metric has useful properties (symmetry, non-negativity, triangular), it has the following drawbacks: 
(i)~it increases when the dimension of the data increases; (ii)~it does not handle cases when data points contain missing values at some dimensions, (iii)~it gives essentially a spatial distance, and does not take into account the positivity of data. For these reasons, following the works of Chang \cite{Chan00:ieeeit}, we prefer information divergences as measures of similarity.

First, each data point is associated with a (positive) normalized quantity:
$$
\tilde{x_i}=\frac{x_i}{\sum_{j=1}^L x_j}
$$
Let $\tilde X =\{ \tilde x_1, \ldots, \tilde x_L \}$ and $\tilde Y$ be defined accordingly.

In the case of our applications, $\tilde X$ (resp. $\tilde Y$) can be interpreted as the probability distribution that a certain amount of information has been measured at wavelength $\lambda_i$ (more precisely around that wavelength). 

The goal is to measure the similarity between these two probability density functions.
Though our aim is to define a similarity measure between two feature vectors, the divergences used will be symmetrized. The same weight will be applied in both directional divergences to the symmetrized one \cite{Bass89:sp}.
$$d(X,Y)=d(X||Y)+d(Y||X)$$
Choosing the popular symmetrized Kullback-Leibler divergence \cite{CoveT91:wiley} leads to : 
 \begin{equation}
d_{KL}(X,Y) = \sum_{i=1}^{L} (\tilde{x_i}-{\tilde y_i})\log \frac{ \tilde{x_i}}{ \tilde{y_i}}. \label{equ:KL}  
 \end{equation}
 Though there exist a wide family of informational divergences, for example Csisz\`ar or Bregman, we have decided to focus our attention only on two popular measures: the Kullback-Leibler and the R\'enyi divergences.
The Kullback-Leibler divergence corresponds to the relative entropy of $\tilde Y$ with respect to $\tilde X$. This information divergence is widely used in information theory, and is strongly related to the Shannon theory \cite{CoveT91:wiley}. 

Alternatively, the symmetrized R\'enyi divergence of order $\alpha$ ($0<\alpha<1$) can similarly be used as a spectral measure: 
\begin{equation}
d_{\alpha}(X,Y)= \frac{1}{\alpha-1}\left(\log{\sum_{i=1}^{L}  \tilde{x_i}^{\alpha}   \tilde{y_i}^{1-\alpha}}  + \log{\sum_{i=1}^{L}  \tilde{y_i}^{\alpha}   \tilde{x_i}^{1-\alpha}}\right).
\label{equ:Renyi}
\end{equation}
Properties and advantages of the R\'enyi $\alpha$-divergence have been detailed by Hero et al. \cite{HeroMMG02:ieeesp}. 
Note that when $\alpha$ tends to $1$, the $\alpha$-divergence (\ref{equ:Renyi}) converges to the Kullback-Leibler divergence (\ref{equ:KL}).
When $\alpha = 1/2$, the $\alpha$-divergence coincides with the Hellinger affinity $d_{1/2}(X\|Y)$ or Hellinger distance, which is often used  to assess how close a probability density is to a reference one.
As the Bayes optimal exponential rate of decay of the decision error in a binary test ('is this spectrum almost identical to this other one?') involves the $\alpha$-divergence of order $1/2$ \cite{HeroMMG02:michigan}, only the value $\alpha=1/2$ will be considered in the following.

Note that these metrics do not satisfy the triangular inequality; therefore they are called semi-metrics. 

In the remote sensing community and in the analysis of similarity measures on hyper-spectral data, a popular metric is the Spectral Angle Mapper (SAM) \cite{Kesh04:ieeegrs}.
\begin{equation}
\theta(X,Y)= \arccos \left( \frac{\langle X,Y \rangle}{\|X\| \|Y\|}\right),
\end{equation}
where $\langle.,.\rangle$ is the dot product, and $\| . \|$ is the Euclidean norm.
This measure is used to define the angle existing between some spectrum and a reference one. It is often called \emph{coefficient of correlation}. It measures a similarity rather than a distance or a dissimilarity. Two objects are considered as close if they are on the same line coming from the origin. The angle $\theta$ belongs to the interval $(0,\pi/2)$.\\

These metrics have been used to determine similarity between points in order to form the MST. In the clustering part of our algorithm, we use the $K$-means algorithm (minimization of the squared error function between the points and the centroids). This classical partitioning algorithm can also be adapted to match the metric used in the construction of the tree. In order to establish a text classification method based on notions of information theory, Dhillon et al. \cite{DhilMK03:jmlr} used a Kullback-Leibler divergence in their objective function  instead of the classical Euclidean distance on the $K$-means algorithm. Banerjee et al. \cite{BaneMDG05:ML} generalized this idea by developing a $K$-means-like algorithm with various distortion functions based on Bregman divergences. In the remote sensing community, Sohn and Rebello \cite{SohnR02:pers} minimized the SAM measure in their objective function.

\section{Results and Discussions}\label{section:5}
In this section, we present some results of the clustering approach obtained through different applications.
The first application is the taxonomic classification of asteroids, by using  reflectance measures at different wavelengths; the second application is the segmentation of multi-spectral satellite images and the third is about the classification of chemical species in hyper-spectral imaging of planet Mars.

\subsection{Asteroid taxonomic study}
Different similarity measures have been used for constructing Prim's MST. We want to show the benefit of using informational divergences to characterize the ``distance" between each pair of points. 

For comparison, clustering results obtained with the so-called \emph{spectral clustering} \cite{ShiM00:ieeepami} approach are included. The spectral clustering algorithm used is that of Ng et al. \cite{NgJW01:nips} based on an eigen-decomposition of the normalized Laplacian of the graph. The affinity that is considered in the latter case relies upon a new metric introduced recently by Grikschat et al. \cite{GrisCHM06:toulouse}, which uses hitting time of Prim's trajectories rooted at each vertex. Some tests have been performed with symmetrical divergences as similarity measures in the spectral clustering algorithm, and are not reported here. On one hand, they have not been proved to be able to handle non Euclidean cases; on the other hand, the results obtained were not better than those reported in Table \ref{SMASSIIresults}. 

The asteroid data are reflectances measured at different wavelengths, from which a mineralogical classification of the asteroid is sought. More details on the physics underlying the classification problems may be found in \cite{MichBR05:toulouse}. 
The observations of the Small Main Belt Asteroid Spectroscopic Survey phase II (SMASSII) was realized between August 1993 and May 1997. These measurements were operated by the  MDM (Michigan Dartmouth MIT) Observatory located in the south west dorsal of Kitt summit in Arizona.

This dataset contains $9$-dimensional spectra of 1341 asteroids taken in the band $[0.44,0.92]$ $\mu m$ and has been used as a reference to the Bus and Binzel taxonomy \cite{BusB02:Icarus}, $25$ classes have been identified from this survey.
Warell and Lagerkvist \cite{WareL07:AA} have established a supervised classification method applied on this dataset. Based on Bus and Binzel taxonomy, the mean spectrum of every class is calculated. Each asteroid is then affected to the closest class (that is, minimum distance with its center), in the meaning of an Euclidean distance. This method is highly dependent on Bus and Binzel taxonomy, which is currently the reference taxonomy used for this survey (that is, this group of asteroids with reflectances measures at the previously given wavelength band). Nevertheless, we have no evidence that this taxonomy will still be valid in the infra-red wavelength for example. This is the major reason why we develop an unsupervised method for clustering this dataset.
To make a fair comparison with \cite{WareL07:AA}, we kept only spectra that do not contain missing values; hence the survey reduces to 1329 asteroids spectra. 

We have also defined functions that allow us to characterize the goodness of grouping.
Though the taxonomy of Bus and Binzel might not be well matched, this is the best available taxonomy for comparison to our results.

A cluster $C$ will be associated with the taxonomic class $Tax$ (defined by either Bus and Binzel or Tholen) which has the largest overlap with $C$. Let us define some variables: $\mathcal{N}_C$ is the number of objects classified in the cluster $C$, $\mathcal{N}_{Tax}$ is the number of asteroids present in the taxonomic class $Tax$ and $\mathcal{N}_{inter}$ corresponds to the number of objects either present in $C$ or in $Tax$. A clustering validity index is defined as $Score = \mathcal{N}_{inter}/\mathcal{N}_{Tax} $. 
$Score$ characterizes the ratio of asteroids belonging to a taxonomic class and that are correctly labeled. 

\begin{table}[h*]
\begin{center}
 \caption{\small{Synthesis of results obtained on SMASSII}}\label{SMASSIIresults}

\begin{tabular}{c|c}
\hline
Clustering Methods & Score \\
\hline
\hline
NN MST (Euclidean) + K-means & 815/1329 = 61,32\% \\
NN MST (Kullback-Leibler) + K-means & 976/1329 = $\mathbf{73,44\%}$ \\
NN MST (Rényi) + K-means & 972/1329 = $\mathbf{73,14\%}$ \\
Spectral Clustering \cite{NgJW01:nips} & 878/1329 = 66,06\%   \\
Spectral Clustering (Dual Rooted Hitting Time) \cite{GrisCHM06:toulouse} & 913/1329 =  68,69\% \\
Warell and Lagerkvist \cite{WareL07:AA} & 777/1329 = 58,46\% \\
K-means (randomly initialized) & 773/1329 = 58,16\% \\
Expectation Maximization (Gaussian Mixture) & 826/1329 = 62,15\%\\
\hline
 \hline
 \end{tabular}

\end{center}
 \end{table}
 
From Table \ref{SMASSIIresults}, we see that properly initialized K-means used simultaneously with informational divergence based affinity measures outperforms previously proposed approaches.\\

\begin{figure*}[htb]
\centering
\subfigure[]{\includegraphics[width=5.5cm,height=5cm]{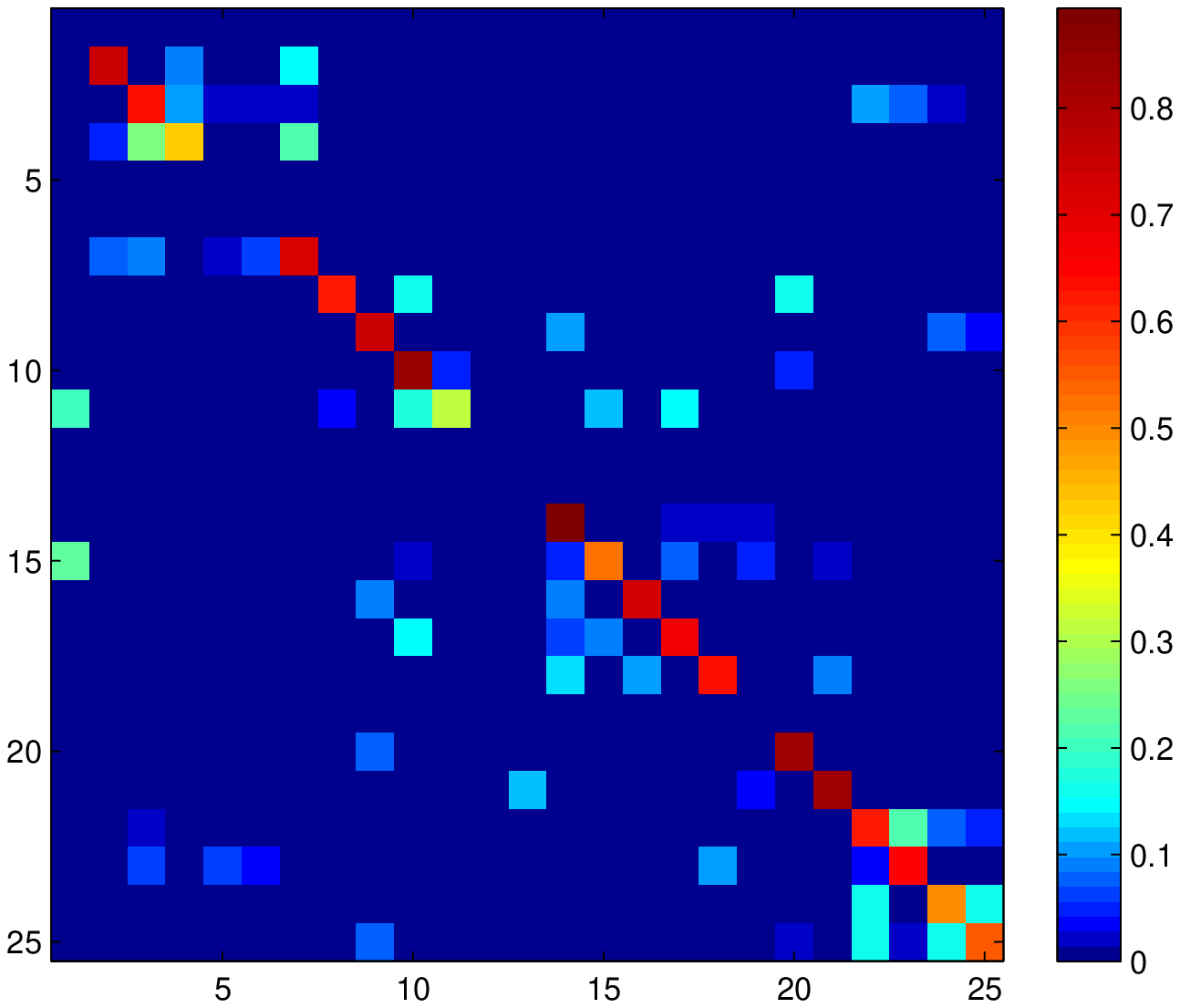}}
\subfigure[]{\includegraphics[width=5.5cm,height=5cm]{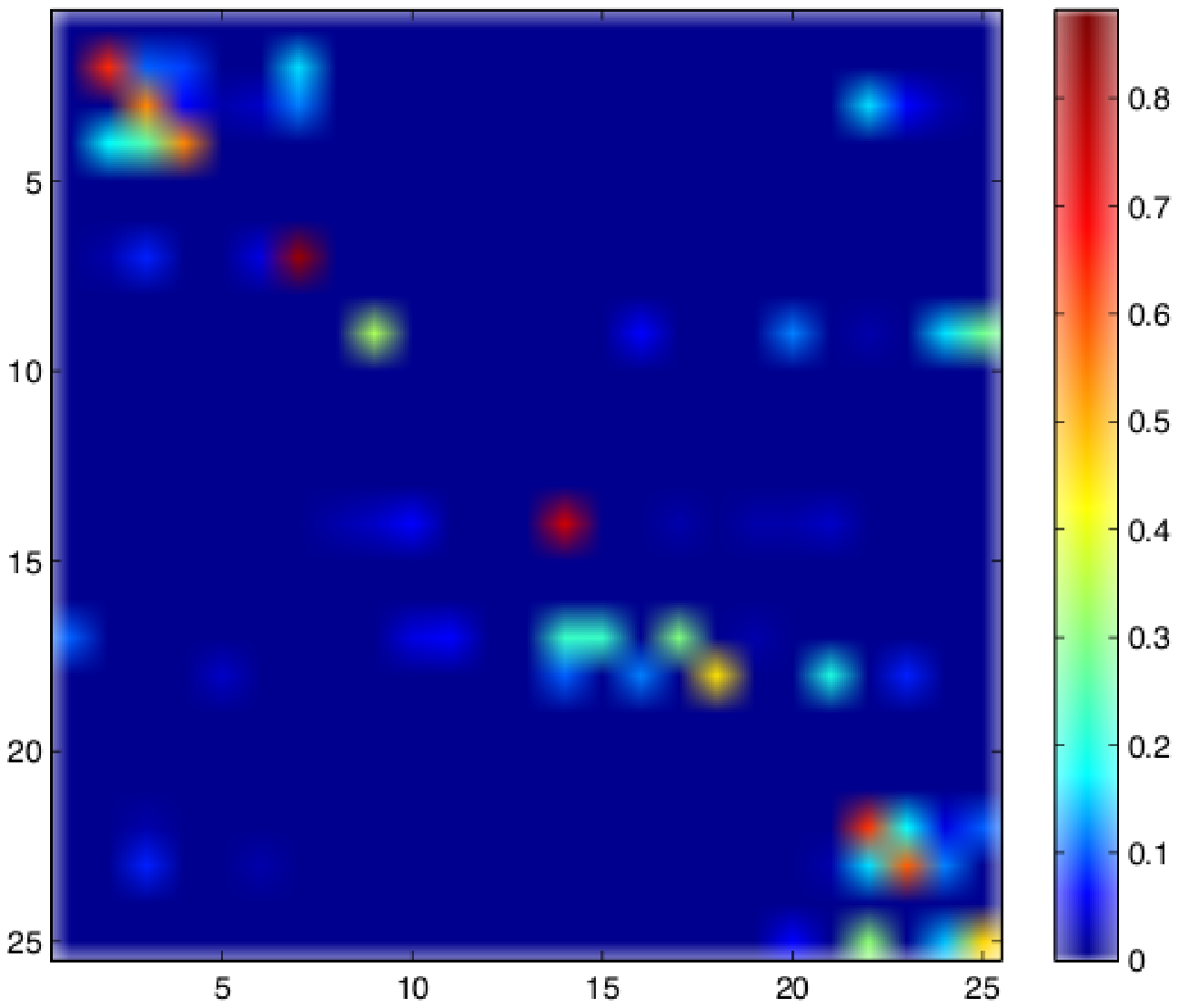}}
\subfigure[]{\includegraphics[width=5.5cm,height=5cm]{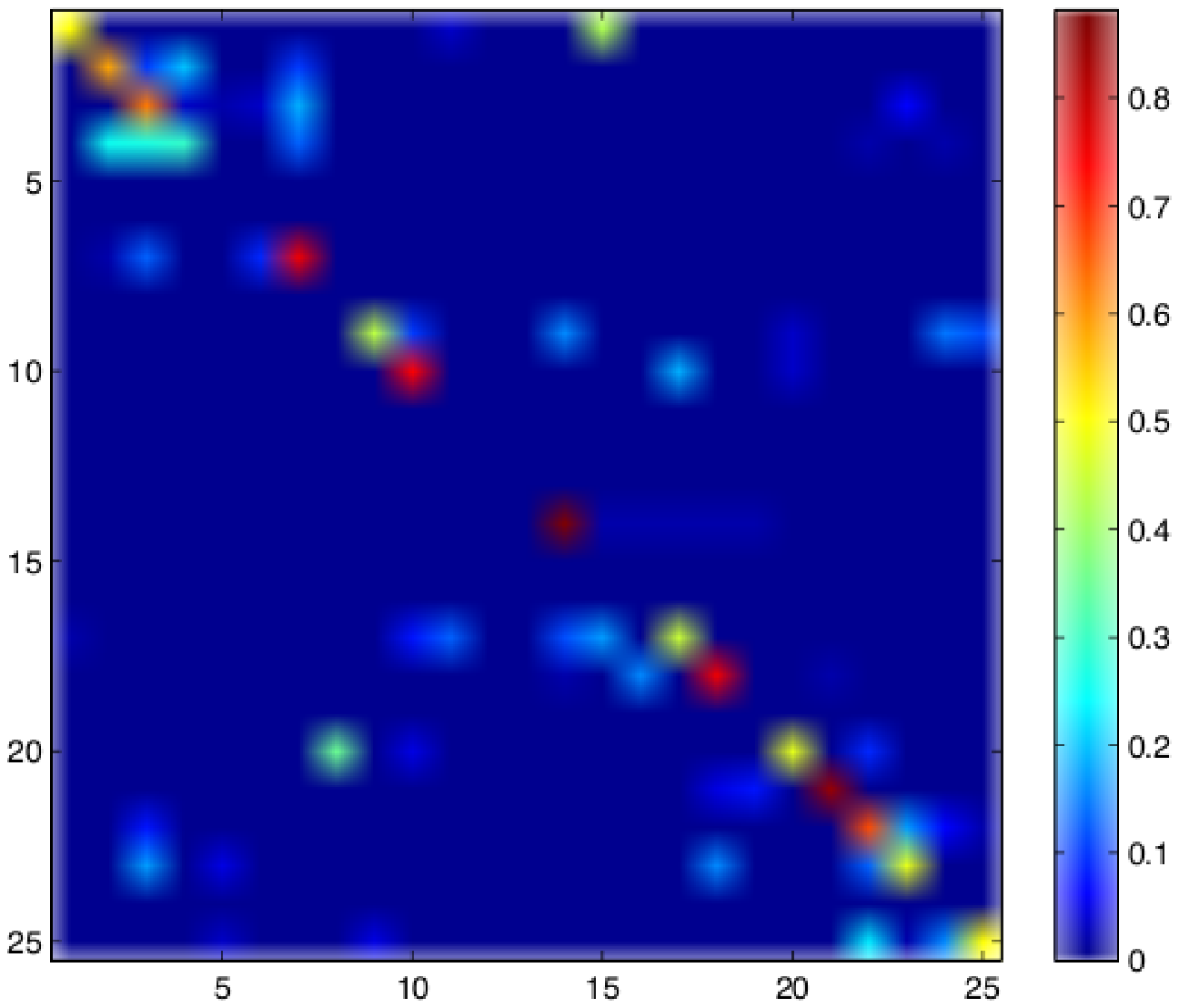}}
\caption{\small{Confusion matrices obtained on the SMASSII data: (a) NN MST (Kullback-Leibler), (b) NN MST (Euclidean) and (c) Spectral Clustering (Euclidean). }}\label{Fig:Confusion}
\end{figure*}

In Fig.~\ref{Fig:Confusion}, we have plotted results for three cases: NN MST with an euclidean distance, NN MST with a Kullback-Leibler divergence and Ng et al. spectral clustering algorithm, in the form of a confusion matrix (also called matching matrix), where each row represents the instances in a predicted class, and each column represents the instances in an actual class. We can notice that not only the use of an information theoretic measure allows us to detect more class but the clusters find better match with the original class than with the other clustering algorithms.

\subsection{Multi-spectral image of Paris}

\begin{figure}[htb]
\centering
\includegraphics[width=8cm,height=5.5cm]{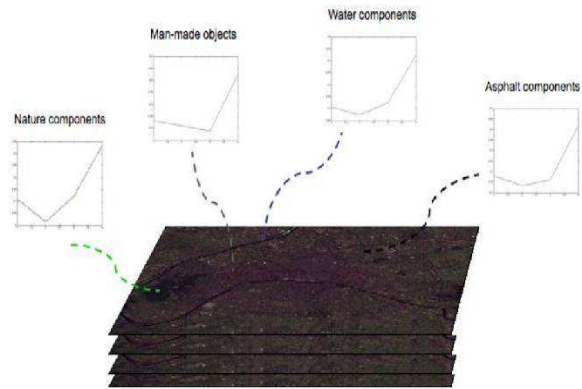}
\caption{\small{Multi-spectral image of Paris composed by multi-components}} \label{Parisoriginal}
\end{figure}

The next set of results has been obtained with a Landsat Thematic Mapper (four-bands) image of size $512 \times 512$ with $30$ meter resolution. Each image is recorded from a device operating at a different wavelength. The images cover areas around the city of Paris (France). Image registration problems are not considered here, and it is supposed that the images are perfectly registered. Fig.~\ref{Parisoriginal} illustrates this. 

In order to compare various clustering results obtained on this multi-spectral image with the similarity measures detailed Section \ref{section:4}, and since we don't have at our disposal any ground-truth classification, we introduce a well-known clustering validity index that does not need reference results: the ratio of intra and inter-cluster variances \cite{DaviB79:ieeepami}. From now on, this index will be called for convenience the Davies Bouldin index.
Denote $\delta_{ij}$ the inter-cluster cross-variance between clusters $i$ and $j$, and $s_i$ the intra-cluster variance within cluster $i$. Then the similarity measure $R_{ij}$ is defined as follows:
$$ 
R_{ij} = \frac{s_i + s_j}{\delta_{ij}}$$
$ \mbox{with } \delta_{ij}=d(\mu_i,\mu_j), ~\mbox{and}~ s_i= \frac{1}{|C_i|} \sum_{v \in C_i} d(v,\mu_i) 
$.\\
The objective to be minimized takes the form
\begin{equation}
DB(C)=\frac{1}{k}\sum_{l=1}^{k}R_l, \quad \mbox{where}~
 R_l= \underset{j\neq i}{\underset{j=1,\ldots ,k}{\max}}  R_{ij}\label{equ:DB}
\end{equation}
It can be clearly seen in (\ref{equ:DB}) that the lower the value the better the partition. 

\begin{table}[htb]
\begin{center}
\caption{{\small Clustering validity indices obtained on the clustering of the multi-spectral image}\label{Paris_results}}
\begin{tabular}{c|c}
\hline
Clustering Methods & Davies-Bouldin index\\
\hline
\hline
NN MST (Euclidean) + K-means & 111.77 \\
NN MST (SAM) + K-means & 157.27 \\
NN MST (Kullback-Leibler) + K-means & $\mathbf{74.87}$\\
NN MST (Rényi) + K-means &  $\mathbf{88.02}$\\
Expectation Maximization (Mixture of Gaussians) & 160.19 \\
K-means (randomly initialized) & 155.13 \\
\hline
\hline
\end{tabular}
\end{center}
 \end{table}
 \medskip 
 
The affinity measure is based upon the Kullback-Leibler divergence between the 4-point spectrum associated with each pixel. The proposed algorithm (Prim based initializing of K-means clustering method) is tested on this multi-spectral image, where each vertex is nothing but a 4-point spectrum. In order to avoid dealing with $512^2 $ vertices, the Nearest Neighbor MST algorithm depicted in Section \ref{section:3} is applied to an image which has been sub-sampled by a factor of 4. 

\begin{figure*}[htb]
\centering
      \subfigure[]{\includegraphics[width=5cm,height=5cm]{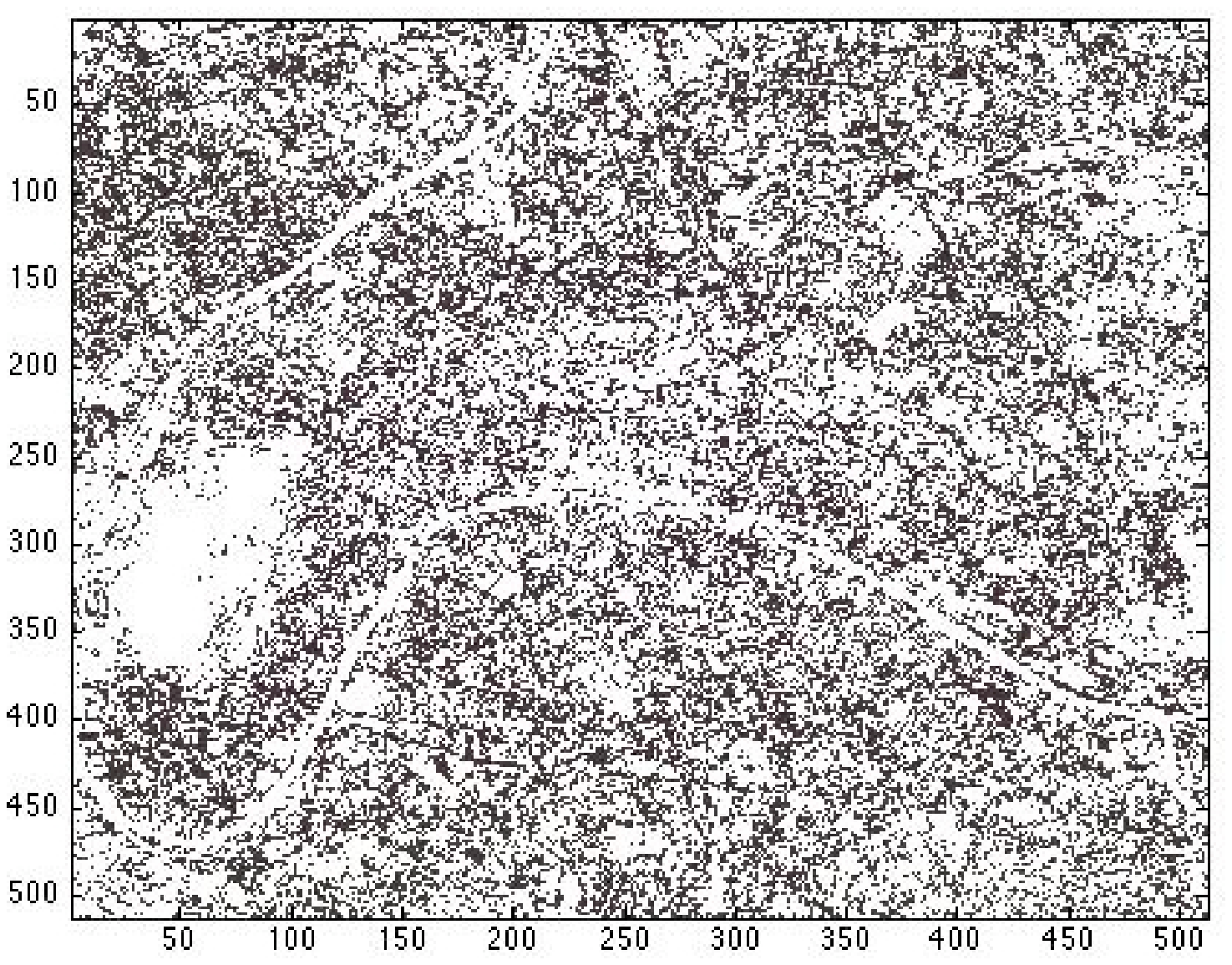}} 
      \subfigure[]{\includegraphics[width=5cm,height=5cm]{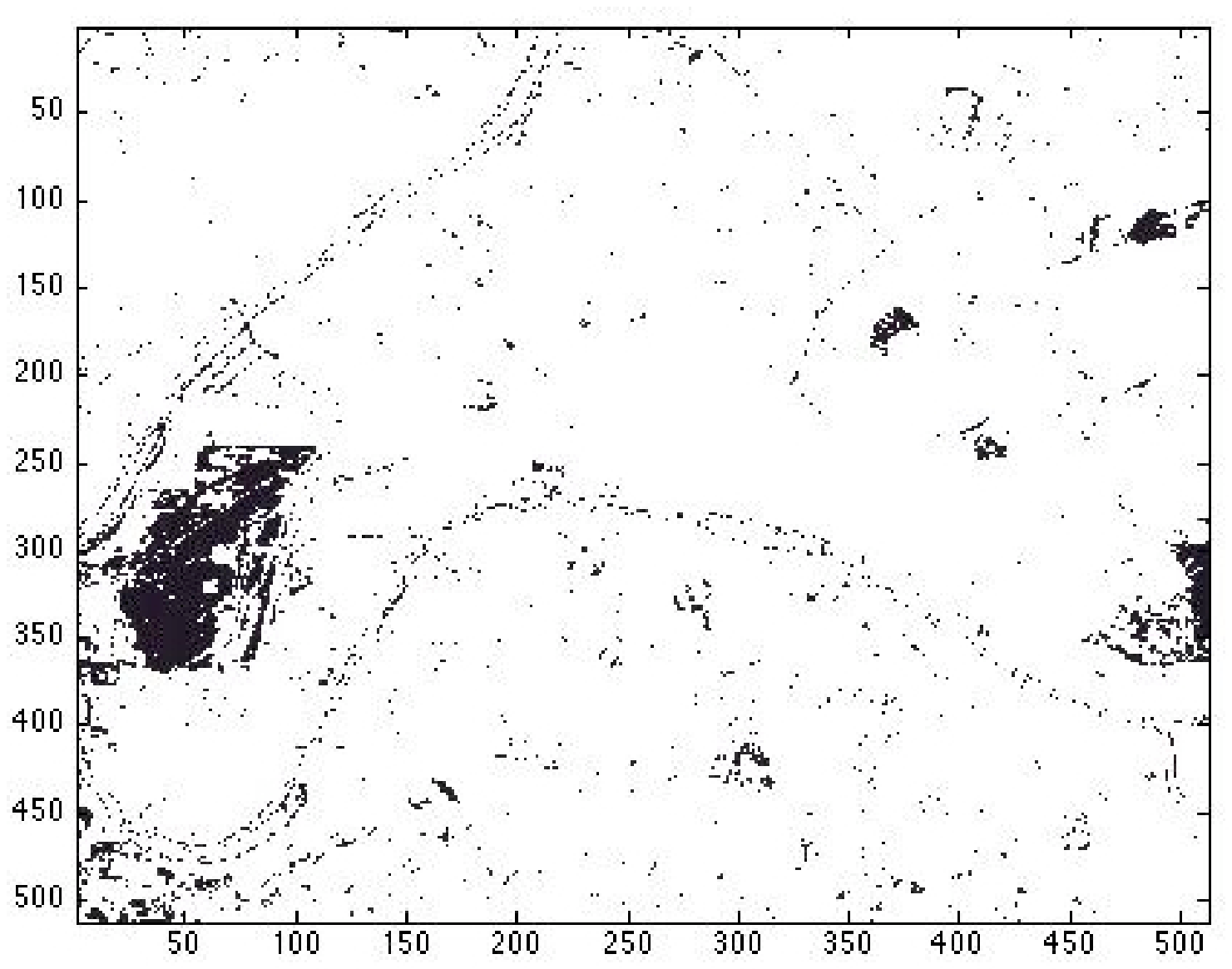}}
      \subfigure[]{\includegraphics[width=5cm,height=5cm]{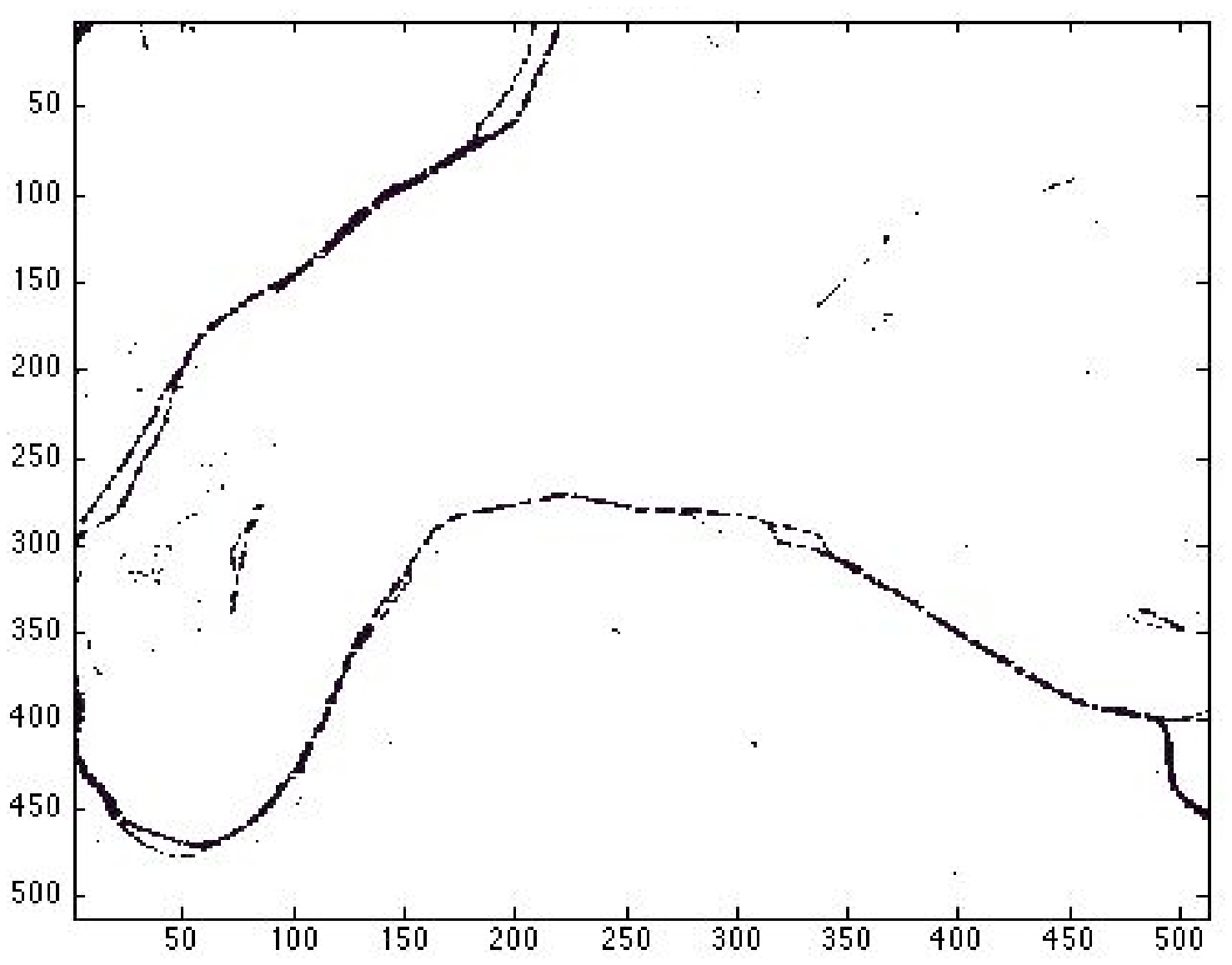}}
      \subfigure[]{\includegraphics[width=5cm,height=5cm]{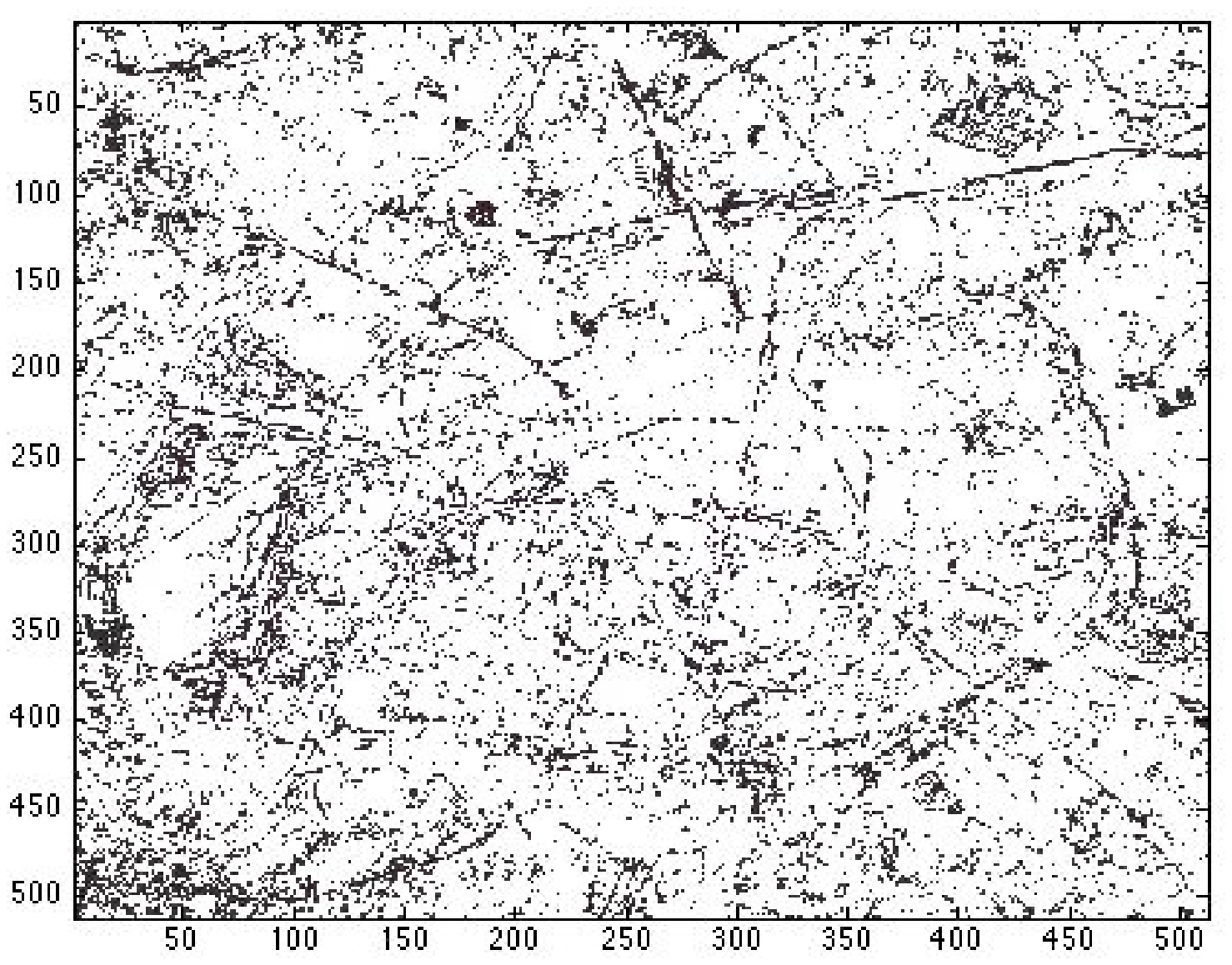}}
      \subfigure[]{\includegraphics[width=5cm,height=5cm]{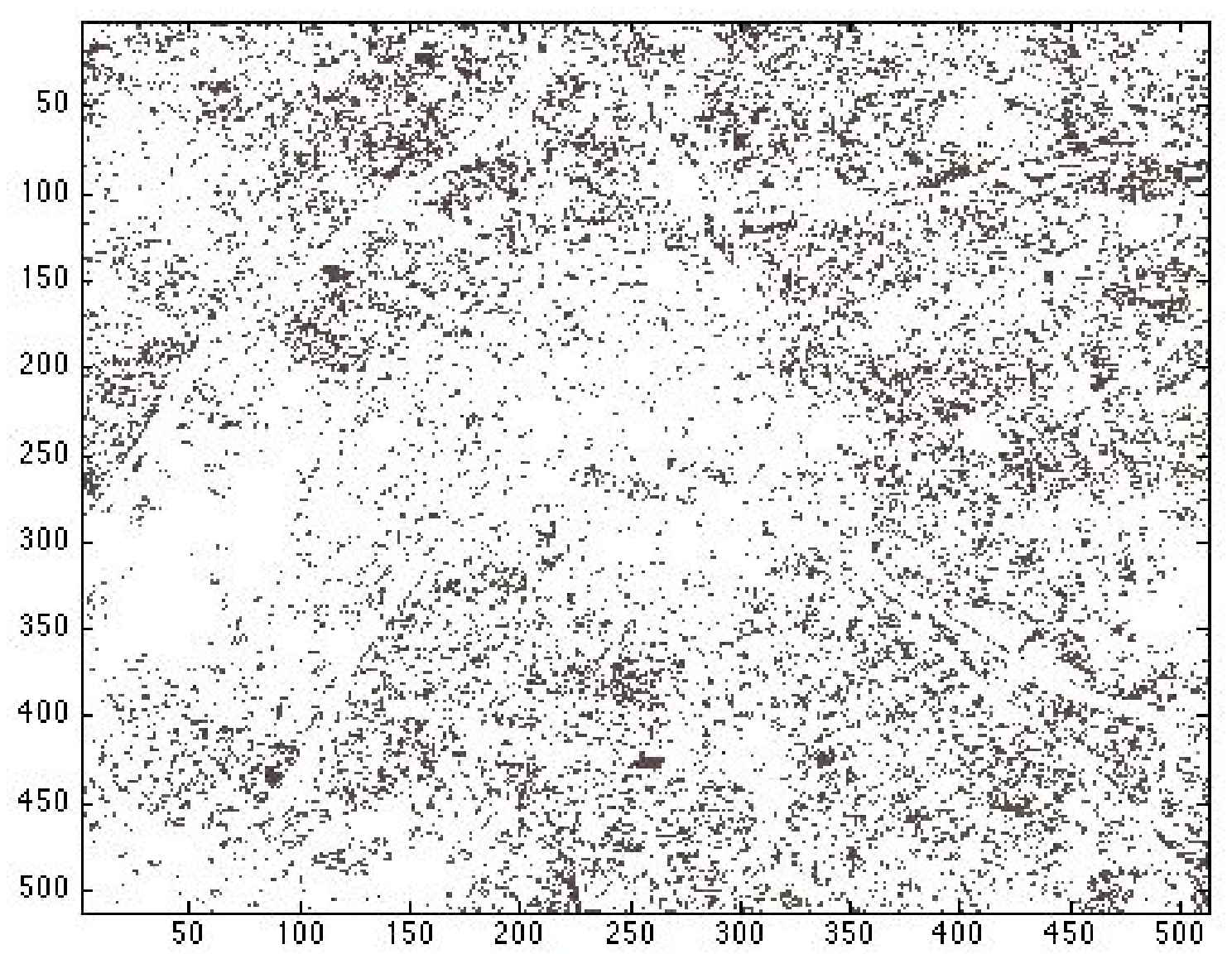}}
      \subfigure[]{\includegraphics[width=5cm,height=5cm]{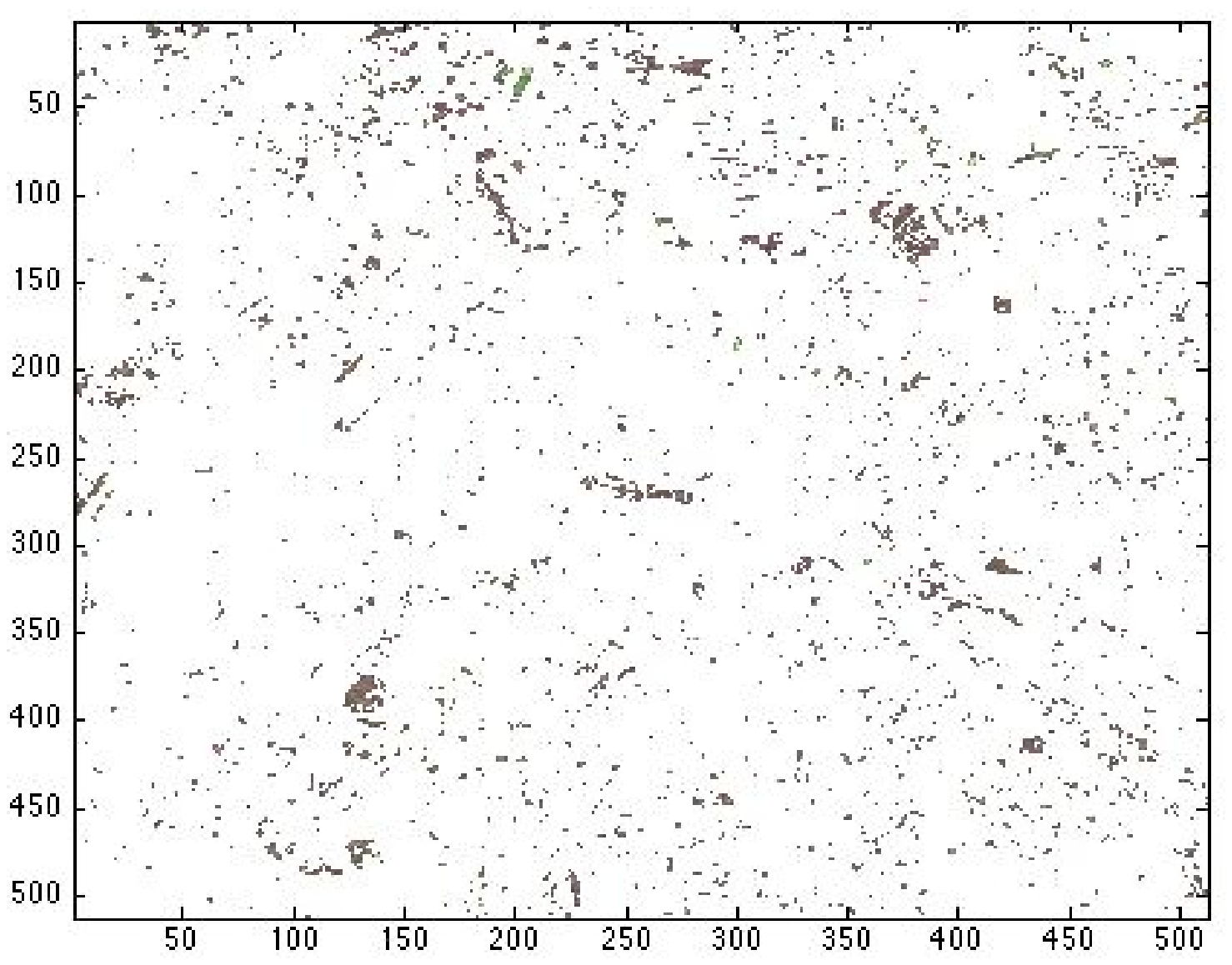}}
       \subfigure[]{\includegraphics[width=5cm,height=5cm]{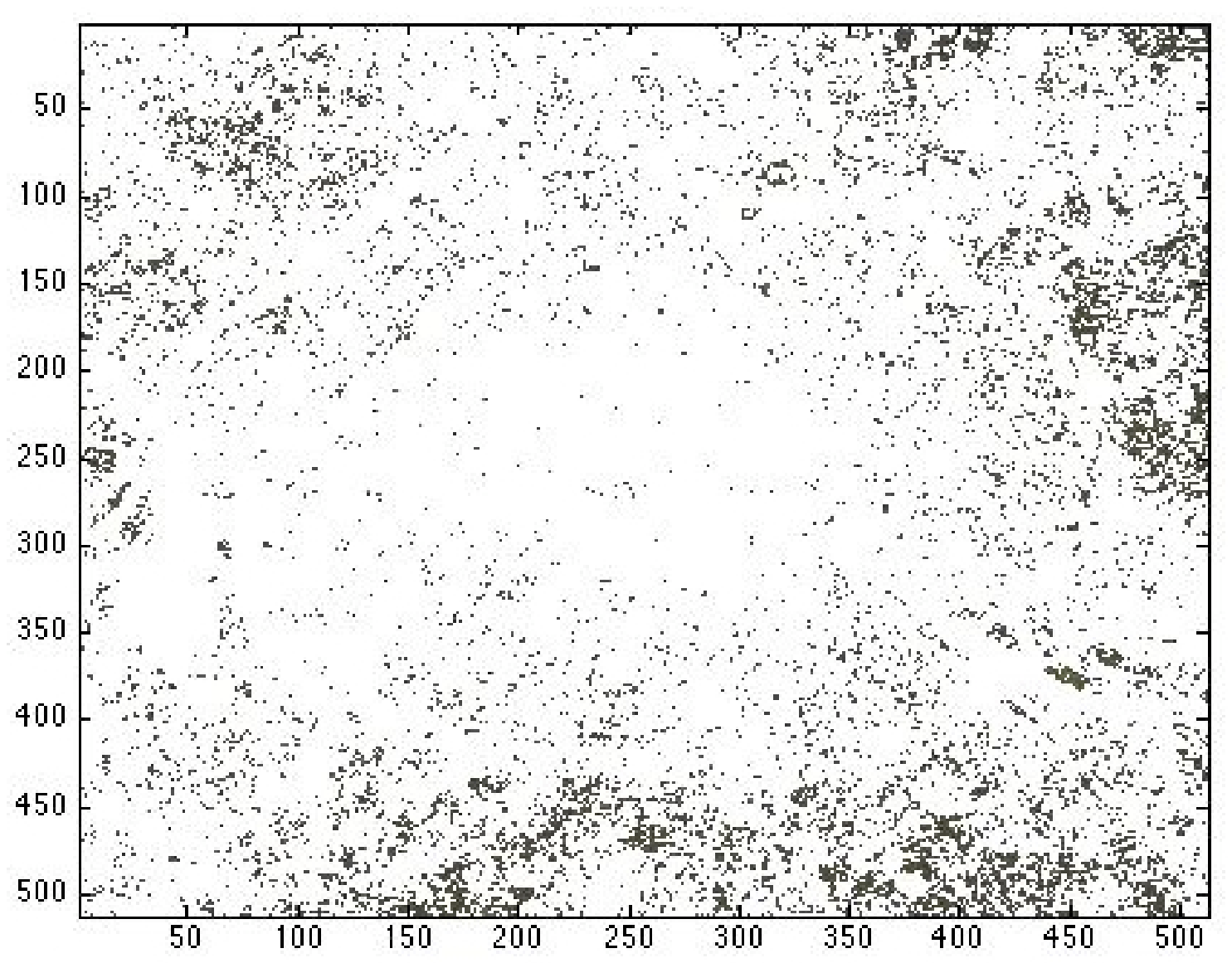}}
      \subfigure[]{\includegraphics[width=5cm,height=5cm]{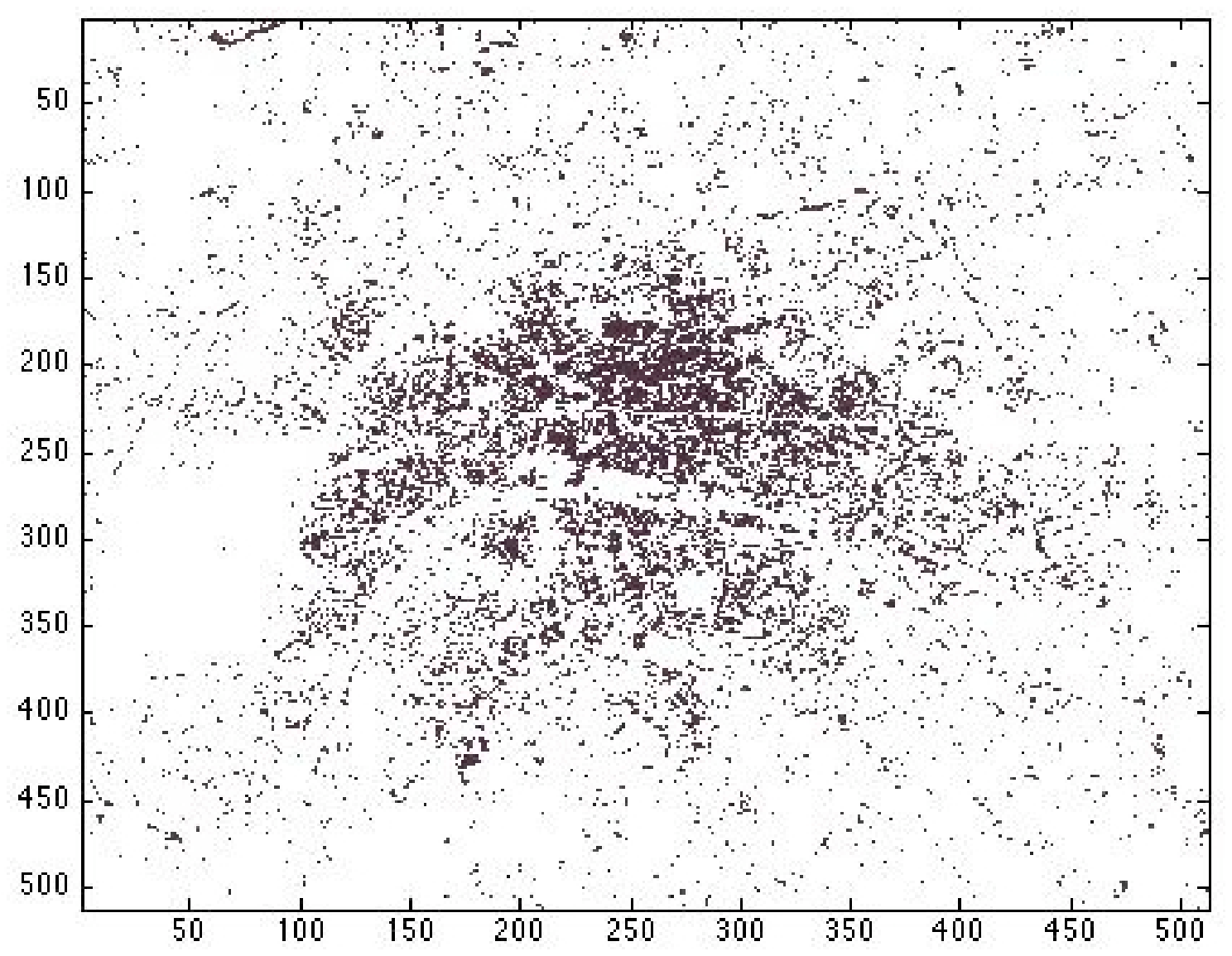}}
\caption{\small{Clusters obtained on the multi-spectral image of Paris: (a) cluster 1, (b) cluster 2, (c) cluster 3, (d) cluster 4, (e) cluster 5, (f) cluster 6, (g) cluster 7 and (h) cluster 8.}}\label{ClustersParis}
\end{figure*}

Fig.~\ref{ClustersParis} shows the obtained results: 8 clusters are identified, from which 3 are easy to understand. Cluster 2 contains the pixels that are characteristic of trees and grass regions. Therefore, one can recognize recreation areas and natural parks in Paris surroundings (Boulogne, Vincennes). Cluster 3 exhibits the 'water areas' in Paris, and the Seine river together with some known ponds is easily extracted.  Cluster 4 is clearly associated with roads, asphalt and concrete. Other clusters cannot be fairly interpreted without cross analyzing our results with for example pollution imaging or gas detection systems.
In Table~\ref{Paris_results}, the Davies Bouldin index is measured on different clustering methods. Our proposed approach with the use of informational divergences gives the best results.

\subsection{Mars hyper-spectral image}
For investigating the efficiency of clustering in high dimension (that is, high number of wavelengths), we present in this section results on a Mars hyper-spectral image. 
This image was provided by the Mars Express (European Space Angency) on board imaging spectrometer instrument OMEGA (Observatoire pour la Minéralogie, l'Eau, les Glaces et l'Activit\'e).  It covers the south polar cap of Mars in the local summer. We are interested in information on this area  taken on two infrared channels: 128 spectral planes from $0.93$ $\mu m$ to $2.73$ $\mu m$ with a resolution of $0.013$ $\mu m$ and 128 spectral planes from $2.55$ $\mu m$ to $5.11$ $\mu m$ with a resolution of $0.02$ $\mu m$. 
Hence, the image is of size $300 \times 120$ pixels in $256$ wavelengths. 
Some pre-treatments have been realized in this image by the `Laboratoire de Planetologie de Grenoble' in order to remove values at certain wavelengths because of faulty pixels of the CCD camera, to calibrate values etc.

\medskip

In the recent years, two methods have been introduced to classify chemical species on this hyper-spectral image of planet Mars. 

The first method developed by Schmidt et al. \cite{SchmDS07:ieeegrs} is actually a supervised approach based on wavelet theory referred to as wavanglet. This method requires \emph{a priori} information, that is, number of classes,  and reference spectra (generated from physical model of chemicals compounds expected or added by the user). The classification is made by computing the spectral angle in the best subspace of a wavelet filtered space. 
The results obtained with this method consist in three binary masks (a pixel belongs to one class) of $CO_2$ ice, $H_2O$ ice and dust. Nevertheless, some pixels have not been classified. A major constraint of the wavanglet classification method is that it is highly dependent on reference spectra, and hence requires an \emph{a priori} report of expert in order to firstly identify all physical compounds present in the scene, and secondly generate (or extract from the image) reference spectra.

The second recent method developed by Moussaoui et al. \cite{MousHSJCBDB08:NC} is based on blind source separation theory on a Bayesian framework referred to as Bayesian positive source separation (BPSS). Various assumptions have been done. The source signals  are assumed to be non-negative (the same holds for the mixing coefficient) and mutually statistically independent. All sources are assumed to be mutually statistically independent and identically distributed, and to follow a Gamma distribution. The noise is assumed to be Gaussian, with zero mean. 

The estimation of source signals, mixing coefficients and hyperparameters (parameters of Gamma distributions and variance of the noise) is carried out using posterior mean estimator and Markov Chain Monte Carlo (MCMC) methods.
Due to the huge computational burden induced by the BPSS method, the authors propose to select a few relevant pixels (with the highest spatial SNR loss) in each relevant independent component provided by a spatial independent component analysis (ICA) algorithm. The number of independent components was estimated from a principal component analysis (PCA) and an \emph{a posteriori} interpretation of the relevant independent components was realized in order to classify only pixels identified as chemical species and not independent components introduced by the imaging spectrometer. The results obtained are not binary masks but abundance fractions of chemical species in this area of Mars. 

\begin{figure*}[htb]
\centering
	\subfigure[]{\includegraphics[width=4.5cm,height=4cm]{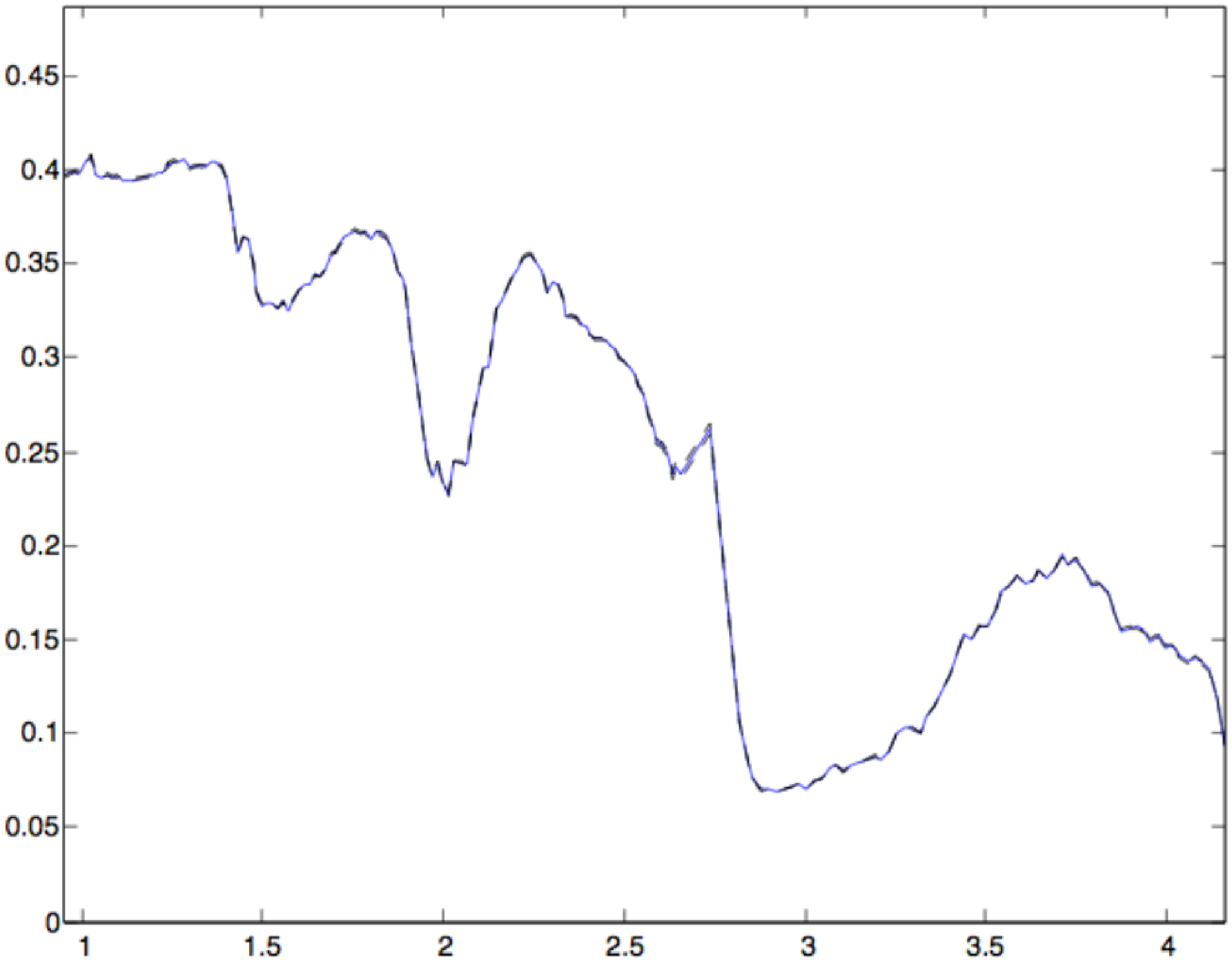}}
	\subfigure[]{\includegraphics[width=4.5cm,height=4cm]{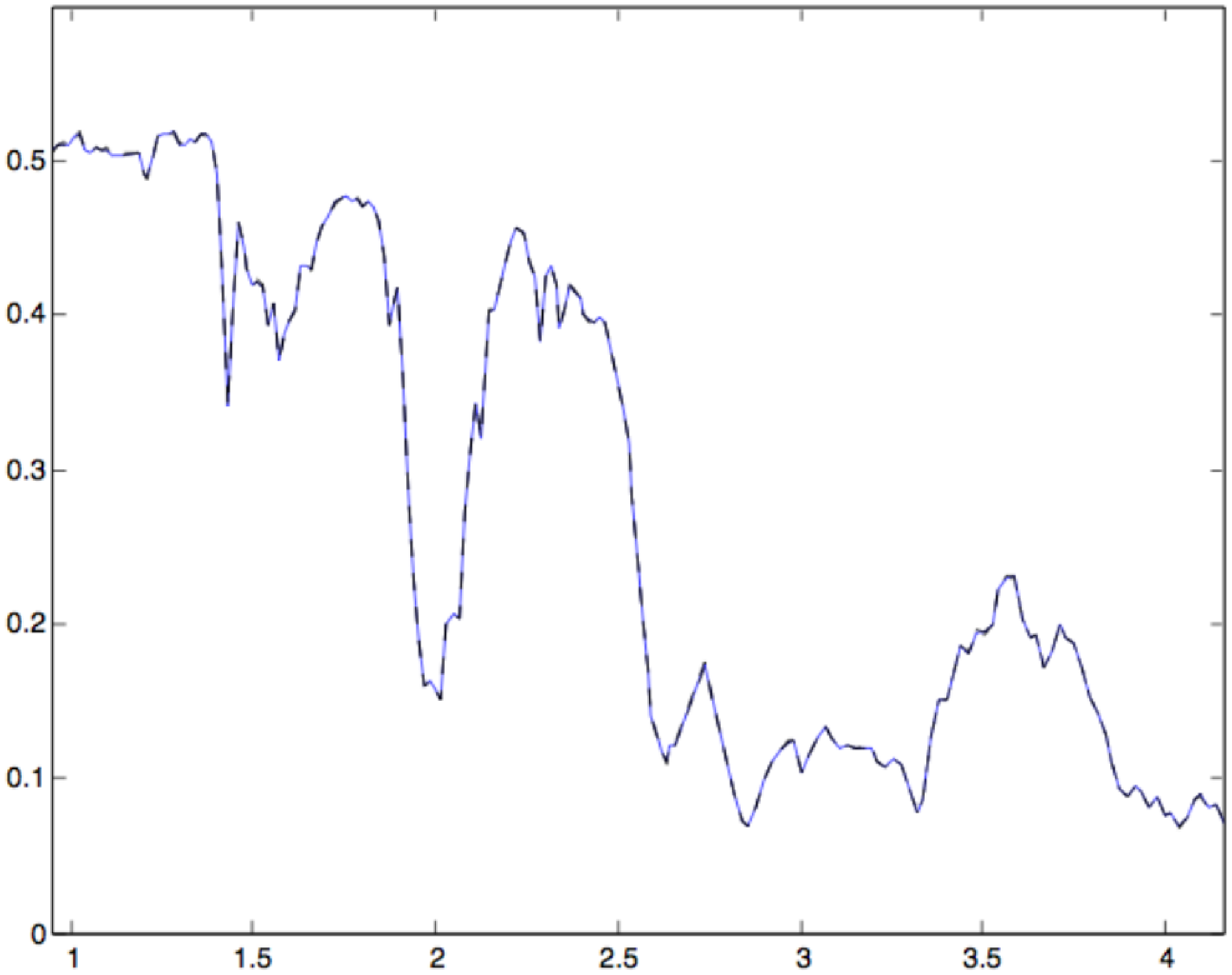}}
	\subfigure[]{\includegraphics[width=4.5cm,height=4cm]{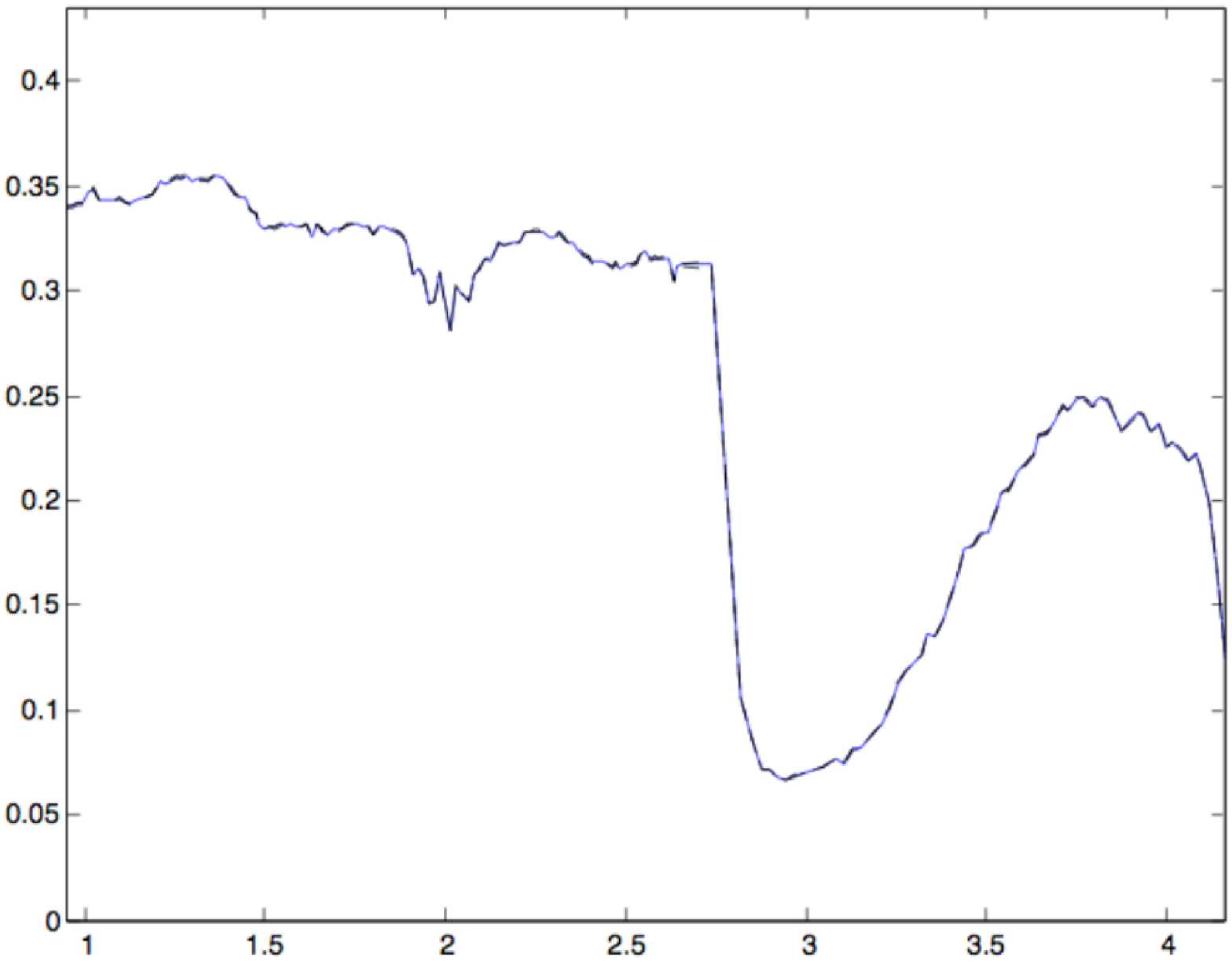}}

 	\subfigure[]{\includegraphics[width=4.5cm,height=6.5cm]{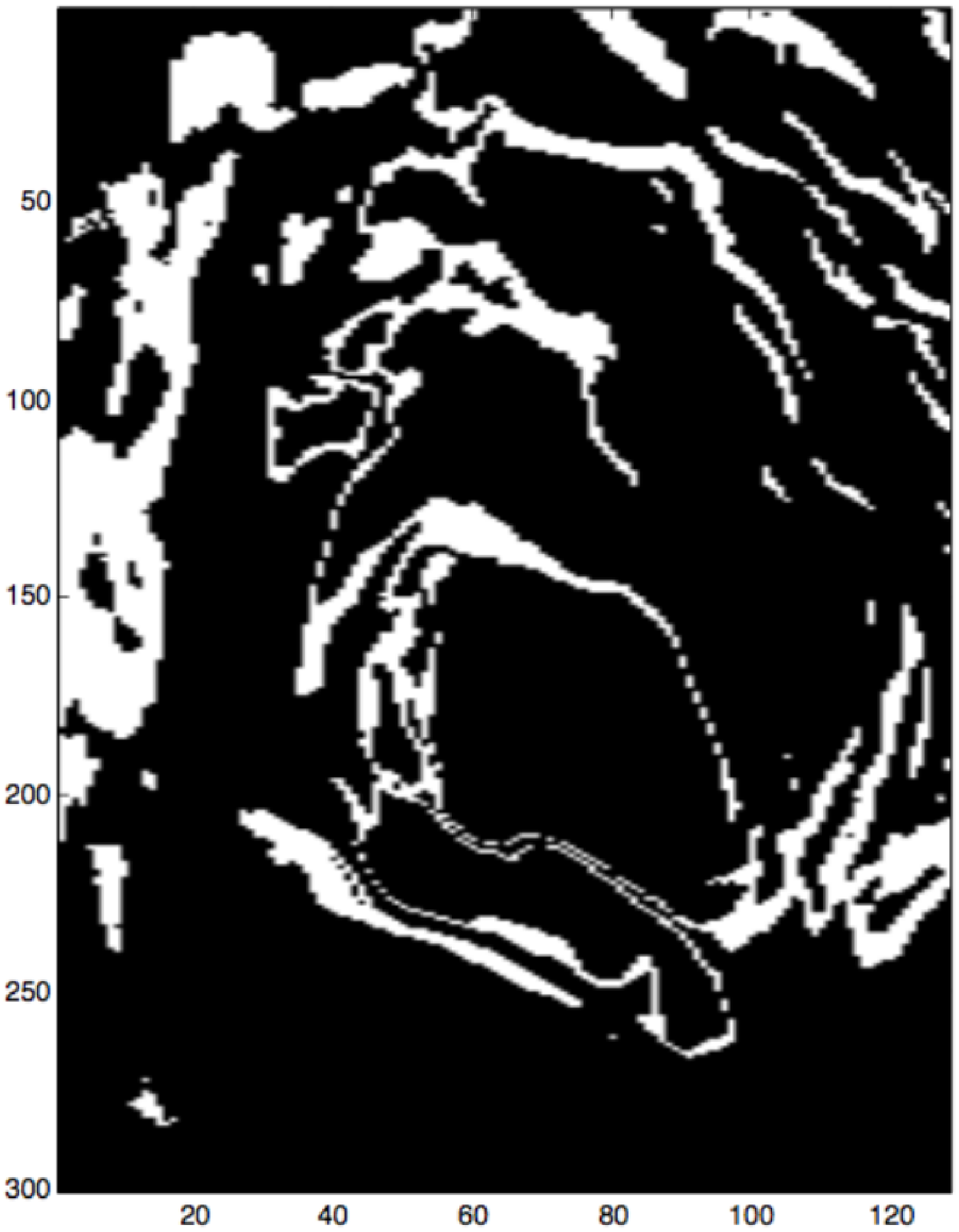}}
 	\subfigure[]{\includegraphics[width=4.5cm,height=6.5cm]{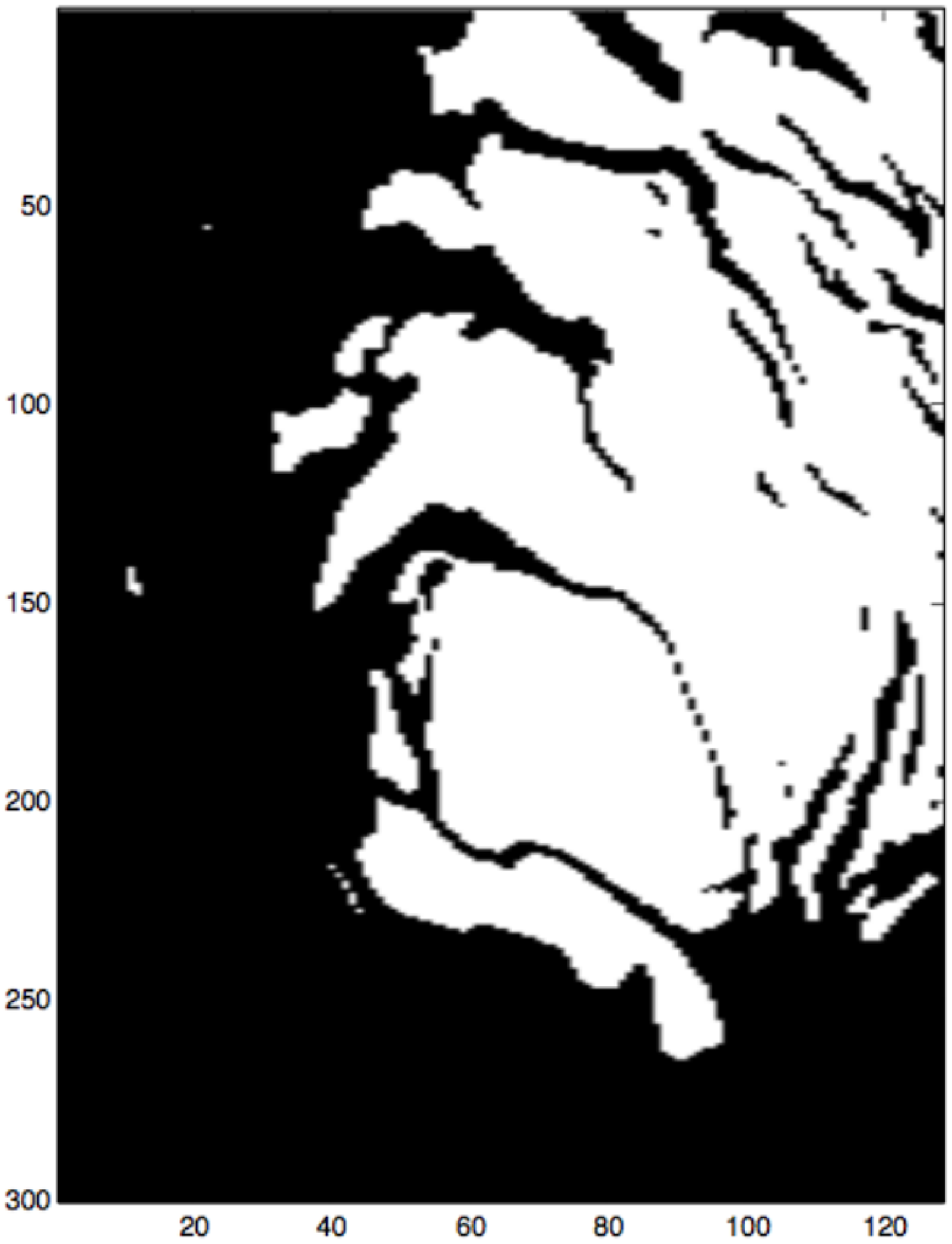}}
\subfigure[]{\includegraphics[width=4.5cm,height=6.5cm]{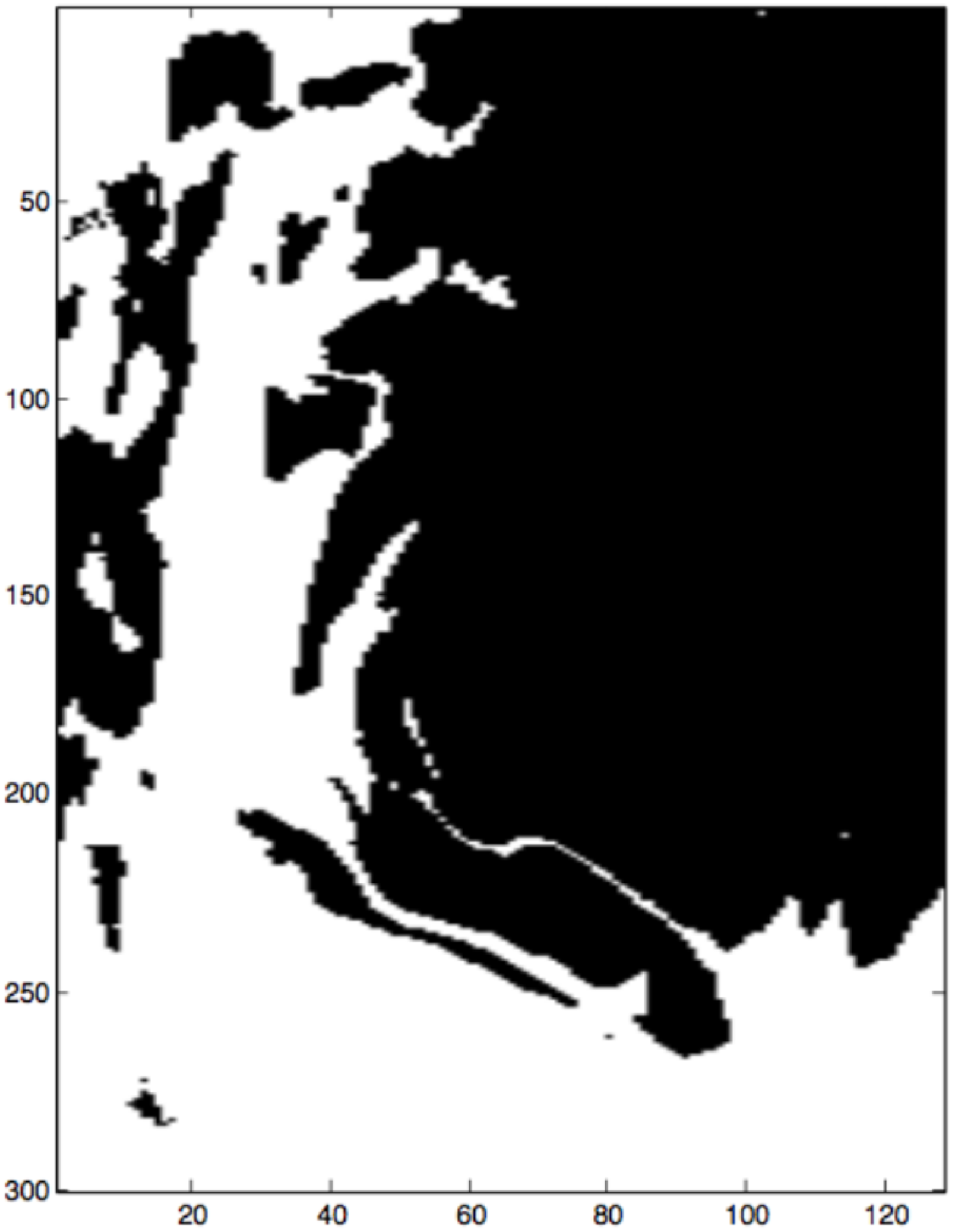}}

\caption{ \small{Mean and $±$ standard deviation spectra obtained: (a)$H_2O$ ice, (b) $CO_2$ ice, (c) dust. Clusters obtained with our method: (d) $H_2O$ ice, (e) $CO_2$ ice and (f) dust.} }
\label{ClustersMars}
\end{figure*}

In Fig.~\ref{ClustersMars}, clusters obtained with our initialization of K-means with information extracted from a Prim's MST are presented. The dissimilarity measure chosen is the Kullback-Leibler divergence. The image presents few homogeneous groups, so the computation of centroids doesn't require the whole image. We have resized the image by column and line sub-sampling (one line over 5 and one column over 3 are kept). The corresponding mean spectrum can also be seen in Fig.~\ref{ClustersMars}. 
The three clusters obtained correspond to $H_2O$ ice, $CO_2$ ice and dust. The representation of the results consists of binary masks (the pixel is white if it belongs to this cluster). $H_2O$ ice compounds are on the peripheral of $CO_2$ ice compounds, which have some sense since at these places the $CO_2$ sublimates to reveal $H_2O$ ice.
The results presented here are similar to those obtained by Moussaoui et al. \cite{MousHSJCBDB08:NC}.

\section{Conclusions and Future works}
In this paper, we have proposed an original approach for clustering multi-dimensional data. The method is based on the estimation of the number of clusters from the construction of a minimum spanning tree, in order to provide the initialization parameters of the classical $K$-means algorithm.

 New criteria are derived for setting the false alarm rate (power) of a test over the Prim's trajectory associated with a MST built over the set of data. 
 We assumed that the vertices are distributed according to a Poisson distribution, in the absence of additional information. Should prior information be available, this reasoning could be extended to other distributions.
The usefulness of the information divergence based affinity measure is illustrated throughout several examples taken from astrophysical field or multi-spectral image analysis.
In this paper, the threshold value is constant along the Prim's trajectory. We can think of setting up a variable threshold as a function of connected segments.
Some improvement in the understanding of the behavior of Prim's trajectory for vertex distributions exhibiting different modes are under study, and will allow to define clusters and labels directly from the MST, without resorting to K-means for example. 
In the case of hyper-spectral images, the proposed method will also require to be developed onto some lower dimensional subspace. Dimension reduction and its relationship to spectral clustering methods applied to graphs using information divergence or MST-based distances (for example, dual rooted tree distances) should be investigated \cite{BelkN02,BachJ06:ML}.

 \section*{Acknowledgments}
  We thank the OMEGA team at IAS/Orsay for his support with sequencing and data download activities. A. Hero's contribution to this paper was partially supported by the US National Science Foundation grant No. CCF 0830490.

\vskip 0.2in
\bibliographystyle{IEEEtran}
\bibliography{Journal}

\end{document}